\documentclass[final,authoryear,5p,times,twocolumn]{elsarticle}
\usepackage{amssymb}
\usepackage{graphicx}
\usepackage{textcomp}
\usepackage{amsmath}
\usepackage{mathtools}
\usepackage{color}
\usepackage{lineno}
\usepackage{url}
\usepackage{enumerate}
\usepackage{epstopdf}
\usepackage{subcaption}
\usepackage{xfrac}
\usepackage{soul}

\modulolinenumbers[5]

\usepackage{graphicx}
\usepackage{graphics}
\usepackage{multirow}
\usepackage{caption}
\usepackage{subcaption}
\usepackage{gensymb}
\usepackage{float}

\usepackage{multirow}
\usepackage{array}
\usepackage{booktabs}
\usepackage{hhline}
\usepackage{xspace}
\usepackage{breqn}

\usepackage{algorithm}
\usepackage{algpseudocode}
\usepackage{soul}

\newcommand{\comst}[1]{}

\definecolor{orange}{rgb}{1,0.5,0}

\newcommand{\com}[1] {}

\newcommand{\ie}{\textit{i}.\textit{e}.,\@\xspace}

\graphicspath{{./figures/}}
\allowdisplaybreaks
\usepackage{array}
\usepackage{arydshln} 
\usepackage{hyperref}

\journal{arXiv}

\begin{document}

\begin{frontmatter}
		
    \title{Less is More: Surgical Phase Recognition with Less Annotations through Self-Supervised Pre-training of CNN-LSTM Networks}
		
		\author[add1]{Gaurav Yengera}
		\ead{g.yengera@gmail.com}
		\author[add2]{Didier Mutter}
		\author[add2]{Jacques Marescaux}
		\author[add1]{Nicolas Padoy\corref{cor1}}
		\ead{npadoy@unistra.fr}
		
		\cortext[cor1]{Corresponding author at: ICube, c/o IRCAD, 1 Place de l'H\^{o}pital, 67000 Strasbourg, France.}
		
		\address[add1]{ICube, University of Strasbourg, CNRS, IHU Strasbourg, France.}
		\address[add2]{University Hospital of Strasbourg, IRCAD, IHU Strasbourg, France.}

\begin{abstract}
Real-time algorithms for automatically recognizing surgical phases are needed to develop systems that can provide assistance to surgeons, enable better management of operating room (OR) resources and consequently improve safety within the OR. State-of-the-art surgical phase recognition algorithms using laparoscopic videos are based on fully supervised training, completely dependent on manually annotated data. Creation of manual annotations is very expensive as it requires expert knowledge and is highly time-consuming, especially considering the numerous types of existing surgeries and the vast amount of laparoscopic videos available. As a result, scaling up fully supervised surgical phase recognition algorithms to different surgery types is a difficult endeavor. In this work, we present a semi-supervised approach based on self-supervised pre-training - i.e., supervised pre-training where the labels are inherently present in the data - which is less reliant on annotated data. Hence, our proposed approach is more easily scalable to different kinds of surgeries. An additional benefit of self-supervised pre-training is that all available laparoscopic videos can be utilized, ensuring no data remains unexploited. In this work, we propose a new self-supervised pre-training approach designed to predict the remaining surgery duration (RSD) from laparoscopic videos, where the labels are automatically extracted from the time-stamps of the video. The RSD prediction task is used to pre-train a convolutional neural network (CNN) and long short-term memory (LSTM) network in an end-to-end manner. Additionally, we present \textit{EndoN2N}, an end-to-end trained CNN-LSTM model for surgical phase recognition, which is optimized on complete video sequences using an approximate backpropagation through time. We provide an apples-to-apples comparison with the two-step training approach where the CNN and LSTM are trained separately (\textit{EndoLSTM}). We evaluate surgical phase recognition performance on a dataset of 120 Cholecystectomy laparoscopic videos (\textit{Cholec120}) and present the first systematic study of self-supervised pre-training approaches to understand the amount of annotations required for surgical phase recognition. The results show that with our self-supervised pre-training approach, similar or even slightly better surgical phase recognition performance can be obtained with 20 percent fewer manually annotated videos and that with 50 percent fewer annotated videos the difference in performance remains within 5 percent. Interestingly, the RSD pre-training approach leads to performance improvement even when all the training data is manually annotated and outperforms the single pre-training approach for surgical phase recognition presently published in the literature. It was also observed that end-to-end training of CNN-LSTM networks boosts surgical phase recognition performance. 
\end{abstract}
		
\end{frontmatter}
	
\section{Introduction}

Surgical phase recognition is an important step for analyzing and optimizing surgical workflow and has been an important area of research within the computer-assisted interventions (CAI) community. Real-time surgical phase recognition technology is essential for developing context-aware systems, which can be used to provide automatic notifications regarding the progress of surgeries and can also be used for alerting the surgeon in the case of an inconsistency in the surgical workflow. Additionally, context-aware systems are important for human-machine interaction within the OR and find applications in surgical education.

The rise of laparoscopic surgery, in addition to improving the quality of surgery for the patient in terms of recovery, safety and cost, provides a rich source of information in the form of videos. Our approach relies purely on these videos for automatically extracting surgical phase information in real-time and does not utilize other sources of information such as tool usage signals, RFID data or data from other specialized instruments since these are not ubiquitous in laparoscopic procedures.

State-of-the-art surgical phase recognition algorithms using laparoscopic videos have achieved good levels of performance with accuracies greater than 80 percent \citep{thesistwinanda}. However, these algorithms are based on fully supervised learning, which limits their potential impact on the development of context-aware systems, since there is a dearth of manually annotated data. Creating manual annotations is an expensive process in terms of time and personnel. Even though there are a few large datasets of manually annotated data, they are available for only a few types of surgeries and cover just a small fraction of the total laparoscopic videos available. For actual clinical deployment there is a clear need for algorithms which are less reliant on manually annotated data in order to scale up surgical phase recognition to different types of surgeries and to be able to use all available data to obtain optimal performance. An unsupervised algorithm would be the ideal solution, however, no purely unsupervised method for training neural networks to effectively recognize surgical phases solely from the frames of laparoscopic videos is known at present. As a result, some degree of supervised learning is necessary. In this regard, we propose an effective semi-supervised algorithm for tackling the problem.

Previous work on self-supervised pre-training \citep{doersch2017multitask} has demonstrated that neural networks can learn a representation of certain inherent characteristics of data by first being trained to perform an auxiliary task for which labels are generated automatically. We propose to pre-train convolutional neural networks (CNN) and long short-term memory (LSTM) networks on the self-supervised task of predicting remaining surgery duration (RSD) from laparoscopic videos \citep{twinanda2018rsd}. We hypothesize that the progression of time in a surgical procedure is closely related to the phases of the surgery and that the variations in surgical phases often correspond to variations in the duration of surgeries. Hence, a model pre-trained to predict RSD could more easily adapt to surgical phase recognition and generalize better to variations in surgical phases. Additionally, the use of self-supervised learning makes it feasible to pre-train the network on a large number of laparoscopic videos. This could enable the network to generalize better to surgeries involving differences in patient characteristics, surgeon skill levels and surgeon styles. In this work, we modify the architecture and training approach used in \cite{twinanda2018rsd} for RSD prediction in order to make it more suitable for pre-training CNN-LSTM networks for surgical phase recognition. Our results show that the pre-training improves performance on the subsequent supervised surgical phase recognition task. Consequently, similar levels of performance could be obtained with less annotated data.

Despite its importance, very few publications have addressed the topic of semi-supervised surgical phase recognition. \cite{bod2017}, the only prior work that we know of to address this problem, presented a method to pre-train CNNs by predicting the correct temporal order of randomly sampled pairs of frames from laparoscopic videos. The idea is to enable the model to understand the temporal progression of laparoscopic workflow, quite similar to the goal of RSD pre-training. While this method does improve performance, its limitations are that it does not utilize complete video sequences to learn about surgical workflow and only the CNN is pre-trained. With the proposed RSD prediction task, the network is pre-trained on complete laparoscopic video sequences. Furthermore, LSTMs, which are responsible for learning temporal features, are pre-trained alongside CNNs. We believe this to be a more effective approach for learning about the temporal workflow of surgical procedures. The experimental results validate the advantages of our proposed RSD pre-training approach. In this work, we also present the first detailed analysis of the effect of self-supervised pre-training on surgical phase recognition performance when different amounts of annotated laparoscopic videos are available. 

It has become a popular choice to combine recurrent neural networks (RNNs) with CNNs for surgical phase recognition \citep{jin2016,bod2017,thesistwinanda}. \cite{jin2016} and \cite{bod2017} trained CNN-RNN models in an end-to-end manner, which enables better correlation between spatial features extracted by the CNN and the temporal knowledge acquired by the RNN. However, due to the high space complexity of such an approach, the RNN is not unrolled over complete video sequences and is optimized on video segments. \cite{thesistwinanda} optimize their model, which we refer to as \textit{EndoLSTM}, over complete video sequences, which is ideal for capturing long range relationships within the surgical procedure, but achieve this by training the CNN and RNN separately in a two-step process. The aforementioned publications have not provided an apples-to-apples comparison between the two methods of training CNN-RNN networks, which we look to address. We propose a model (\textit{EndoN2N}), which optimizes a CNN-RNN network in an end-to-end manner on complete video sequences using an approximate backpropagation through time (BPTT) and compare it to the \textit{EndoLSTM} model based on the same architecture. Understanding the best method of training surgical phase recognition models is important for obtaining optimal performance. This helps when scaling up surgical phase recognition to different types of surgeries, since a better optimized model will require less annotated data to obtain the required levels of performance. We observe that end-to-end training leads to superior performance and better generalization within the different surgical phases.

The innovation presented in this paper can be summarized as follows: (1) introduction of RSD prediction as a self-supervised pre-training task, which outperforms the previous self-supervised pre-training approach proposed for surgical phase recognition, (2) self-supervised pre-training of CNN-LSTM networks in an end-to-end manner on long duration surgical videos, (3) the first systematic study of semi-supervised surgical phase recognition performance with variation in the amount of annotated data and (4) apples-to-apples comparison between an end-to-end CNN-LSTM training approach (\textit{EndoN2N}) and the two step optimization used in the \textit{EndoLSTM} model. We also present additional experiments to better understand the characteristics of the proposed RSD pre-training model and examine the potential of our models for actual clinical application.
	
\section{Related Work}\label{sec:related_work}
	
	\subsection{Self-Supervised Learning}
	
	Unsupervised representation learning has been an active area of research within the context of deep learning. Initial work on the topic focused on methods for initializing deep neural networks with weights close to a good local optimum, since no method was known at the time to effectively train randomly initialized deep networks. One of the most popular approaches was to learn compact representations which could be used for reconstruction of the input data \citep{hinton504,bengio2006}. \cite{hinton504} demonstrated a method for initializing the weights of a deep autoencoder through unsupervised training of stacked single-layer restricted Boltzmann machines (RBMs). \cite{bengio2006} showed that deep neural networks can be initialized with meaningful weights by training each layer as an individual autoencoder. Both these works highlighted the performance improvement obtained from unsupervised pre-training for subsequent supervised learning tasks.

	The availability of large datasets containing millions of labeled high-resolution images made it possible to effectively train deep CNNs for vision tasks without relying on any form of pre-training. \cite{kri2012alexnet} trained a randomly initialized deep CNN on the ImageNet dataset for image recognition, which considerably outperformed previous state-of-the-art machine learning approaches. Since then, for several computer vision applications \citep{girshick2014, donahue2017}, pre-training CNNs using supervised learning on large datasets, such as ImageNet, has become the norm as it tends to outperform unsupervised pre-training approaches. However, large datasets of annotated data are not available in all domains and are difficult to generate. Hence, unsupervised representation learning is still a very attractive option, especially if it can match the performance of purely supervised pre-training approaches.
	
	Self-supervised learning has been recently introduced as an alternate method for unsupervised pre-training. The goal is to learn underlying relationships in real-world data by utilizing inherent labels. Several approaches were presented to capture visual information from static images which would be beneficial for subsequent supervised learning tasks such as image classification and object detection. \cite{doersch2015} built a siamese network to predict the relative position between randomly sampled pairs of image patches in order to learn spatial context within images. \cite{noroozi2016} extended the method to arrange multiple randomly shuffled image patches in the correct order, essentially making the network solve a jigsaw puzzle. \cite{zhang2016} and \cite{larsson2016} proposed to pre-train CNNs by making them predict the original color of images which have been converted to grayscale. \cite{dosovitskiy2014} created surrogate classes corresponding to single images and extended the classes by applying several transformation to the images. They then pre-trained CNNs by learning to differentiate between different surrogate classes while being invariant to the transformations applied. However, the aforementioned self-supervised pre-training approaches are not ideal for a task such as surgical phase recognition, where it is beneficial to utilize video data rather than static images, as it possesses temporal information in addition to visual information. 
	
	Several works have explored self-supervised representation learning using video data. \cite{mobahi2009} presented a method to learn temporal coherence in videos by enforcing that the features extracted using a CNN from consecutive images be similar. \cite{agarwal2015} utilize egomotion as a supervisory signal for self-supervised pre-training. \cite{wang2015} proposed to learn video representations by tracking image patches through a video. \cite{misra2016unsupervised}, \cite{fernando2017}, \cite{lee2017opn}, and \cite{bod2017} all aimed to learn representations that capture the temporal structure of video data. \cite{misra2016unsupervised} pre-trained CNNs by predicting if a set of frames are in the correct temporal order and they formulated the task as a binary classification problem. \cite{fernando2017} sampled subsequences, containing both correct and incorrect temporal sequences, from videos and trained a network to distinguish the subsequences that have an incorrect temporal order. \cite{lee2017opn} trained a network to sort a sequence of randomly shuffled image frames into the correct temporal sequence. The method proposed by \cite{bod2017} involved predicting the correct order of a pair of frames, which have been randomly sampled from a laparoscopic video, and is very similar to the approach of \cite{misra2016unsupervised}. All of these approaches focus on pre-training CNNs. CNN-LSTM networks are often utilized for applications related to action recognition, where the LSTM is the critical component for learning temporal structure within video data. Hence, we claim that it is not always optimal to merely pre-train the CNN. We believe that pre-training both the CNN and LSTM networks would be ideal for learning representations that capture correlations between the spatial and temporal structure of video data.
	
	For pre-training LSTM networks, future prediction in videos has been proposed as a self-supervised learning task \citep{srivastava2015,lotter2017}. It was argued that a learned representation which could be used to predict future frames of a video, would gather knowledge about temporal and spatial variations. \cite{srivastava2015} train an LSTM encoder-decoder network to simultaneously predict future frames and reconstruct a video sequence. This pre-training approach was shown to improve performance on activity recognition tasks. \cite{lotter2017} present a network for video prediction, comprising of CNNs and Convolutional LSTM networks \citep{shi2015}, inspired by neuroscience research on 'predictive coding'. The network was trained in an end-to-end manner and was shown to be more effective in predicting the future frames of a video as compared to an LSTM encoder-decoder network, but the potential for utilizing this approach as a pre-training step for action recognition was not explored. Despite future prediction approaches being viable for pre-training CNN-LSTM networks, they have only been validated on short video sequences. Our proposed method aims to obtain long-range spatio-temporal knowledge by utilizing complete laparoscopic videos, which are of long durations. 
	
    \begin{figure*}[t]
    \centering
    \includegraphics[width=18cm]{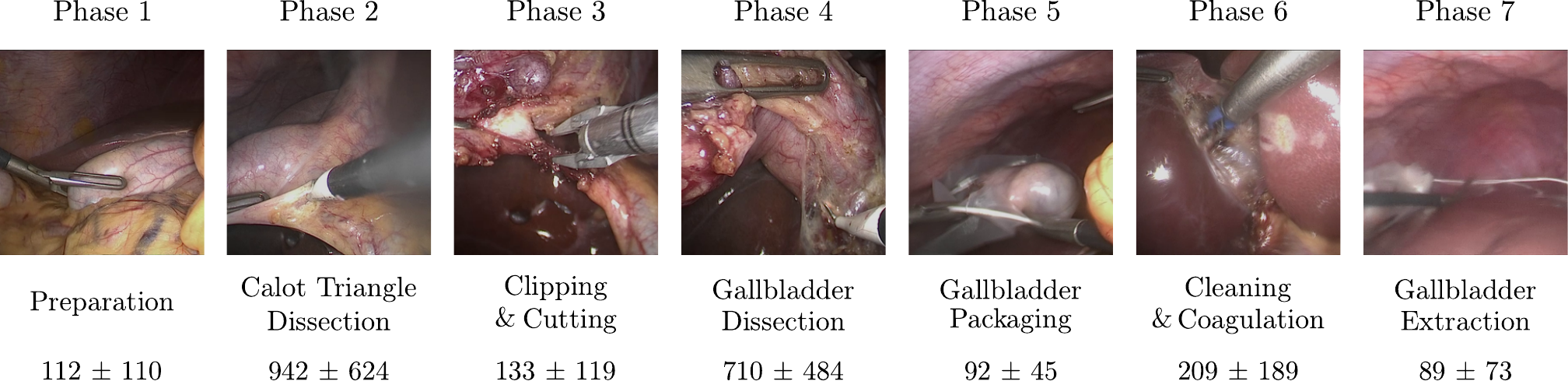}
    \caption{Cholecystectomy surgical phases with mean ($\pm$ std) duration in seconds within the \textit{Cholec120} dataset.}\label{fig:phases}
    \end{figure*}
        
	\subsection{Surgical Phase Recognition}
	
	Previous surgical phase recognition approaches have usually relied on either visual data \citep{twinanda2017endonet, blum2010}, tool usage signals \citep{padoy2012, forestier2013} or surgical action triplets \citep{forestier2015, katic2014}. Among these, visual information is the only source of data that is ubiquitous in all laparoscopic surgical procedures, whereas tool usage and triplet information are obtained either through specialized equipment or manual provision. Since the focus of this work is on the development of real-time surgical phase recognition approaches suitable for widespread deployment in ORs, we propose models which rely purely on visual data, though they could be extended to utilize other data too. In this section, only the previous works which also utilize visual data are discussed.
	
	Various statistical models have been utilized for modeling the temporal structure of surgical videos. Hidden Markov Models (HMMs) were used in \cite{padoy2008}, \cite{lalys2012} and \cite{cadene2016}. \cite{derga2016} implemented a Hidden semi-Markov Model. \cite{twinanda2017endonet} utilized hierarchical HMMs; and Conditional Random Fields have also been a popular choice \citep{quellec2014, charrire2017, lea2015}. Some works have also utilized Dynamic Time Warping \citep{blum2010, lalys2013}, which is not applicable for real-time surgical phase prediction though, as the algorithm requires information from the entire video and is also not well suited for complex non-sequential workflows.
	
	\cite{padoy2008} and \cite{derga2016} combined tool usage signals and visual cues from laparoscopic images for real-time surgical phase recognition, however widespread application of surgical phase recognition algorithms relying on tool usage signals seems to be a difficult task. Several works proposed effective approaches for surgical phase recognition in cataract surgeries \citep{lalys2012, quellec2014, quellec2015, charrire2017}. These approaches relied on handcrafted features though and it was shown in \cite{twinanda2017endonet} that automatically extracting features using a CNN significantly outperformed commonly utilized handcrafted features. 
    The approach proposed by \cite{cadene2016} is to provide a HMM with features extracted using a deep CNN. However, the use of a RNN, such as the LSTM, for temporal sequence learning is shown to perform better than HMMs \citep{thesistwinanda}.
	
	Approaches that combine CNNs with RNNs have been presented in \cite{thesistwinanda}, \cite{jin2016} and \cite{bod2017}. The \textit{EndoLSTM} model presented in \cite{thesistwinanda} utilized a two-step approach of training a CNN and a LSTM independently for surgical phase recognition. The CNN was used to extract features specific to the surgical phases from the frames of laparoscopic videos, which were then provided to the LSTM during training. The LSTM was optimized on complete video sequences. However, theoretically, end-to-end training is ideal for combining the complementary spatial and temporal knowledge captured by the CNN and RNN networks respectively \citep{hajj2017}.
    
    Practically, end-to-end training of a CNN-RNN network on complete laparoscopic video sequences is not feasible due to the high space complexity of the approach. Previous works have presented end-to-end training approaches optimized on video subsequences. \cite{jin2016} performed end-to-end optimization of a CNN-LSTM network over a set of 3 frames sampled at regular intervals from a laparoscopic video. This was the best performing model at the M2CAI 2016 surgical workflow challenge\footnote{http://camma.u-strasbg.fr/m2cai2016/index.php/workflow-challenge-results/}, outperforming the \textit{EndoLSTM} model. However, it is not evident if the performance improvement is due to the alternate training approach or the utilization of a deeper CNN. \cite{bod2017} incorporated a gated recurrent unit (GRU) \citep{cho2014} and trained a CNN-GRU network in an end-to-end manner on laparoscopic video subsequences. They copy the GRU's hidden state between consecutive subsequences belonging to the same video sequence. However, this model was unable to match the performance of \textit{EndoLSTM} on the EndoVis15Workflow\footnote{http://endovissub-workflow.grand-challenge.org/} dataset and the authors attributed this to the large cholecystectomy specific surgical dataset used to train \textit{EndoLSTM}. Essentially, no previous publication has provided an apples-to-apples comparison between an end-to-end optimization approach and the two-step optimization approach of \cite{thesistwinanda}.

    \begin{figure*}[t]
       \begin{subfigure}[t]{0.45\textwidth}
    	\centering
        \includegraphics[width=8.5cm]{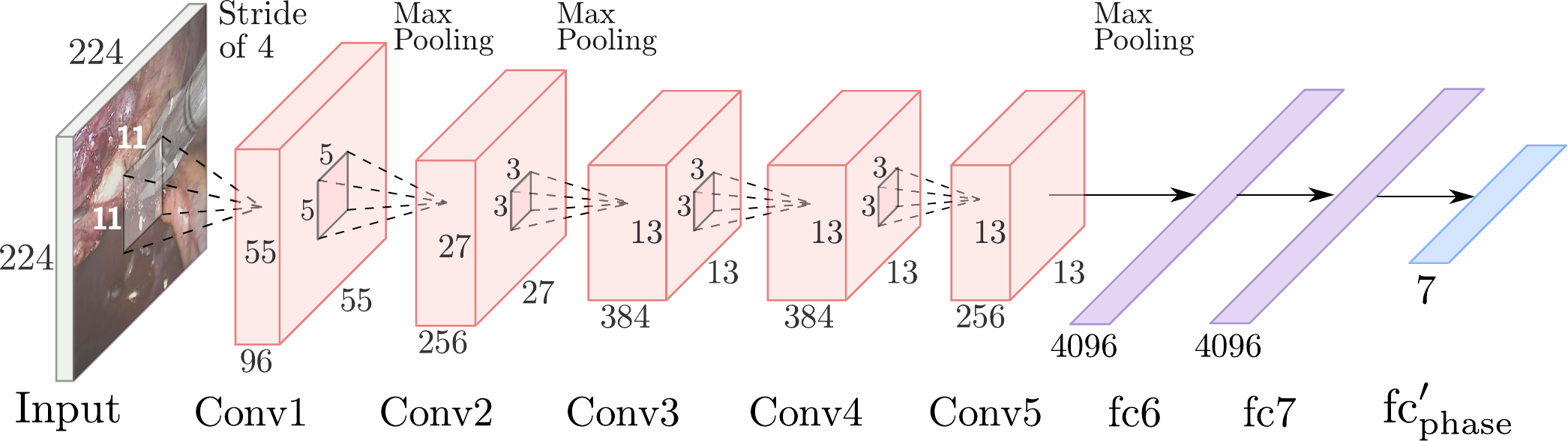}
        \caption{CNN Fine-Tuning Architecture}\label{fig:cnn_finetuning}
        \end{subfigure}
       \begin{subfigure}[t]{0.55\textwidth}
    	\centering
        \includegraphics[width=9cm]{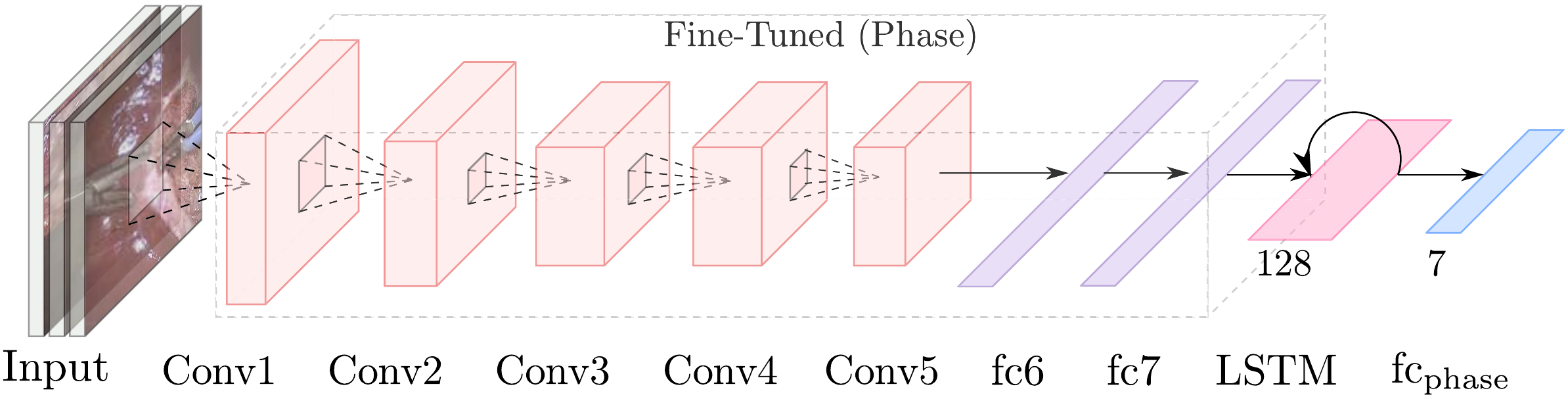}
        \caption{CNN-LSTM Architecture}\label{fig:cnn_lstm}
        \end{subfigure}
        \caption{\textit{EndoN2N} model for surgical phase recognition. The initial CNN fine-tuning network is depicted in (a) and (b) depicts the CNN-LSTM network which is trained in an end-to-end manner, where the layers within the dotted line are initialized by the CNN fine-tuning step.}\label{fig:endon2n}
    \end{figure*}
    
    \section{Methodology}
    
    In this paper, surgical phase recognition approaches for cholecystectomy surgeries are discussed. It is to be noted though that the proposed approaches are generalizable to other surgery types as well. We divide cholecystectomy surgical procedures into 7 distinct surgical phases, depicted in Figure \ref{fig:phases}, similar to \cite{twinanda2017endonet}. We classify each time step of a laparoscopic video as one of the 7 surgical phases, hence formulating the surgical phase recognition task as a multi-class classification problem.
    
    This section will first discuss the CNN-LSTM architecture of the proposed \textit{EndoN2N} model. This will be followed by a detailed presentation of our end-to-end training approach to contrast it with existing CNN-LSTM training approaches. 
    Then, the proposed RSD pre-training will be presented. The motivation for using this pre-training task along with the proposed RSD prediction model and the corresponding surgical phase recognition model will be discussed. Finally, we will briefly describe the temporal context pre-training approach of \cite{bod2017}, which we use as a comparison baseline for our pre-training approach, since it is the only self-supervised pre-training approach previously validated for surgical phase recognition.
    
     \begin{figure}[ht]
		\begin{subfigure}[ht]{0.15\textwidth}
    	\centering
		\includegraphics[height = 3.3cm]{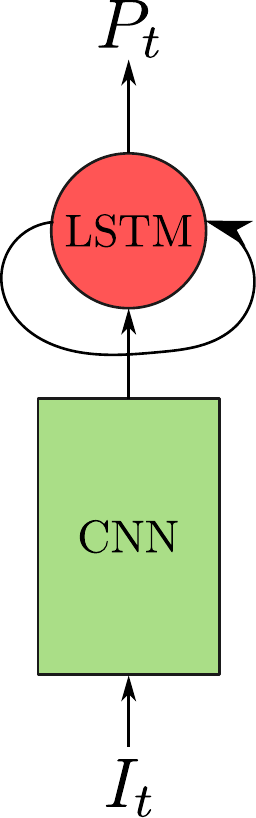}
    	\caption{Rolled graph}\label{fig:rolled}
		\end{subfigure}
        \begin{subfigure}[ht]{0.35\textwidth}
    	\centering
		\includegraphics[height = 3.3cm]{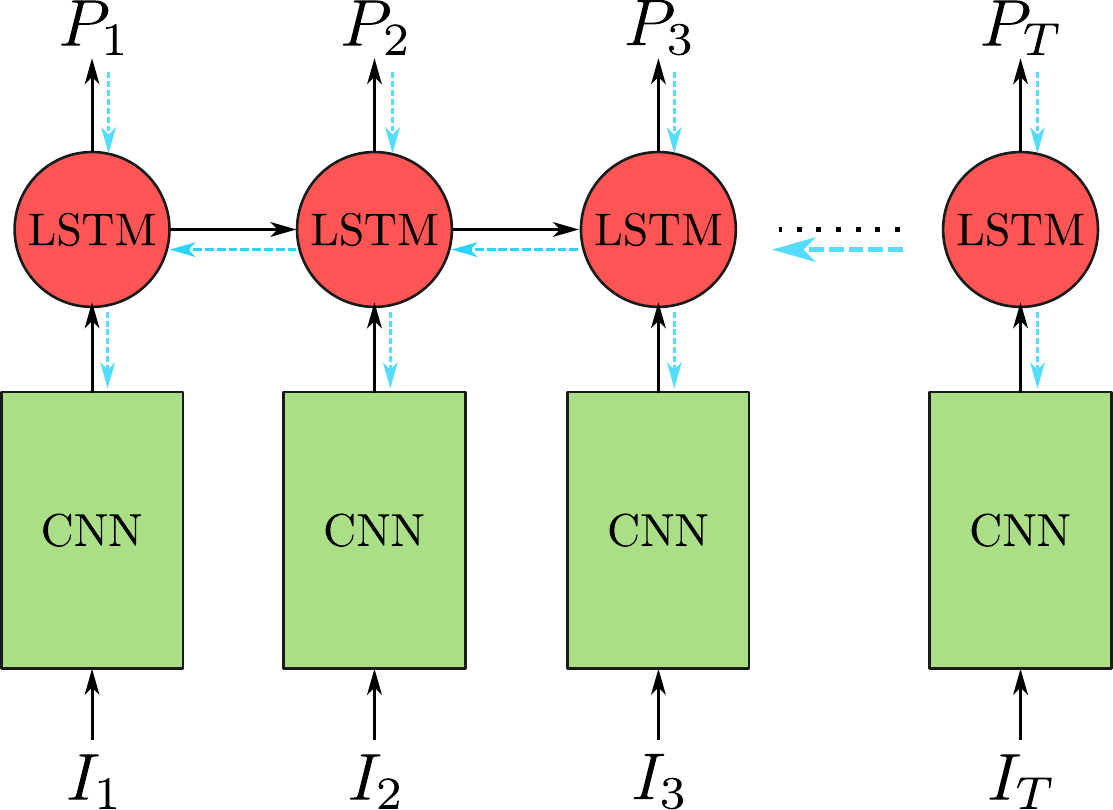}
    	\caption{Unrolled graph}\label{fig:unrolled}
		\end{subfigure}
        \caption{Computation graph for end-to-end training of a CNN-LSTM network. $I_t$ and $P_t$ are the input frame and network prediction at the $t^{\textnormal{th}}$ time-step of a sequence, respectively. (a) shows the rolled CNN-LSTM network. (b) depicts the unrolled computational graph, with the blue lines illustrating the backpropagation of the loss through the CNN-LSTM network for a sequence of length $T$.}\label{fig:graph}
    \end{figure}
    
    \begin{figure*}[t]
       \begin{subfigure}[t]{0.4\textwidth}
    	\centering
        \includegraphics[height=2.3cm]{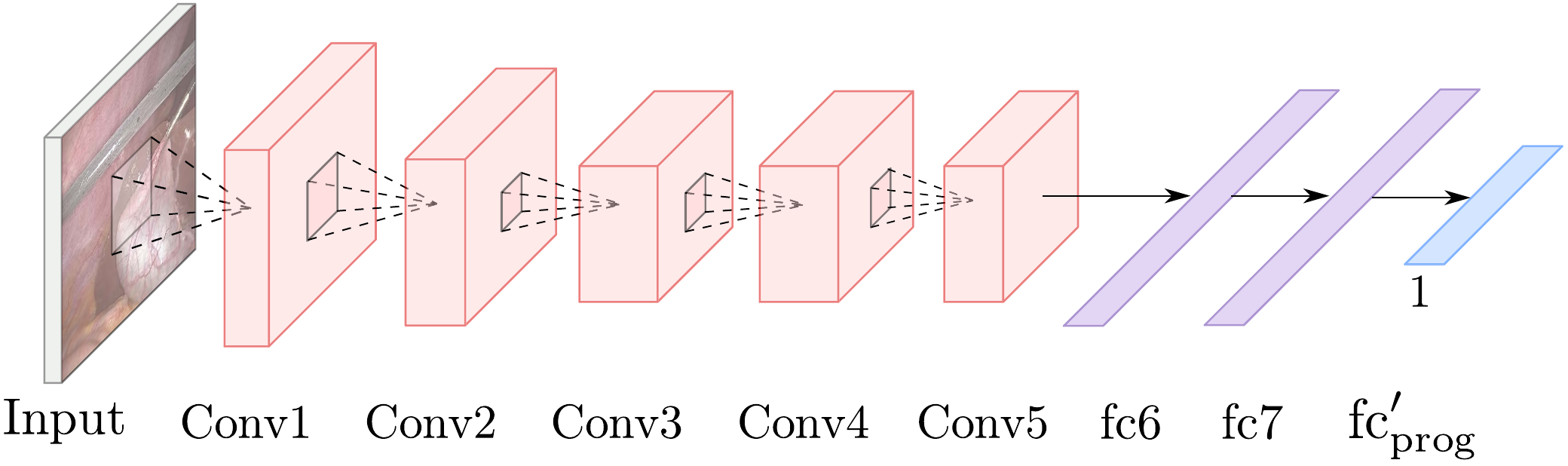}
        \caption{Progress Regression Architecture}\label{fig:progress_cnn}
        \end{subfigure}
       \begin{subfigure}[t]{0.6\textwidth}
    	\centering
        \includegraphics[height=2.3cm]{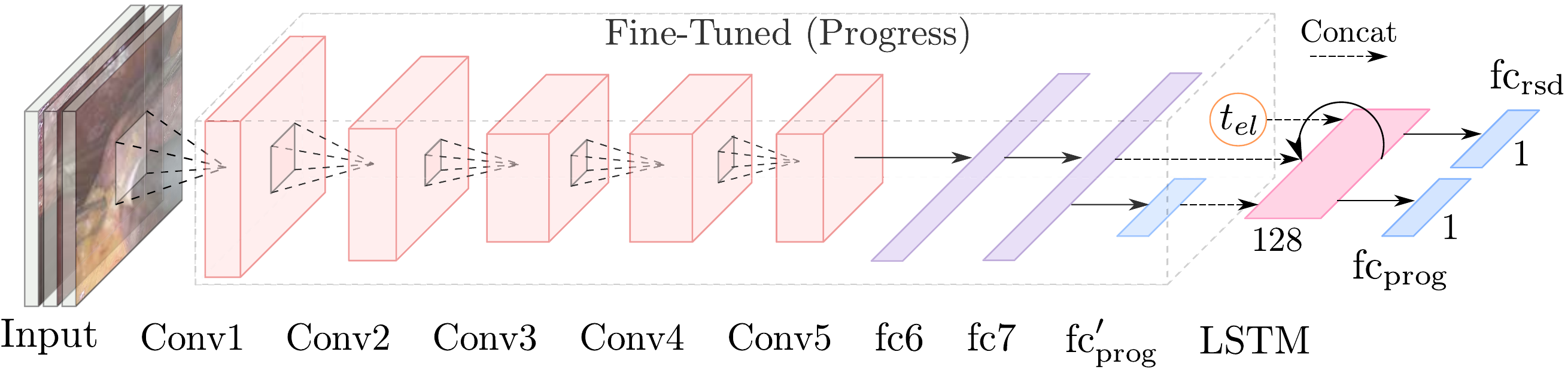}
        \caption{RSD and Progress Multi-Task Network}\label{fig:rsdnet2}
        \end{subfigure}
        \begin{center}
        \begin{subfigure}{0.5\textwidth}
        \centering
        \includegraphics[height=2.3cm]{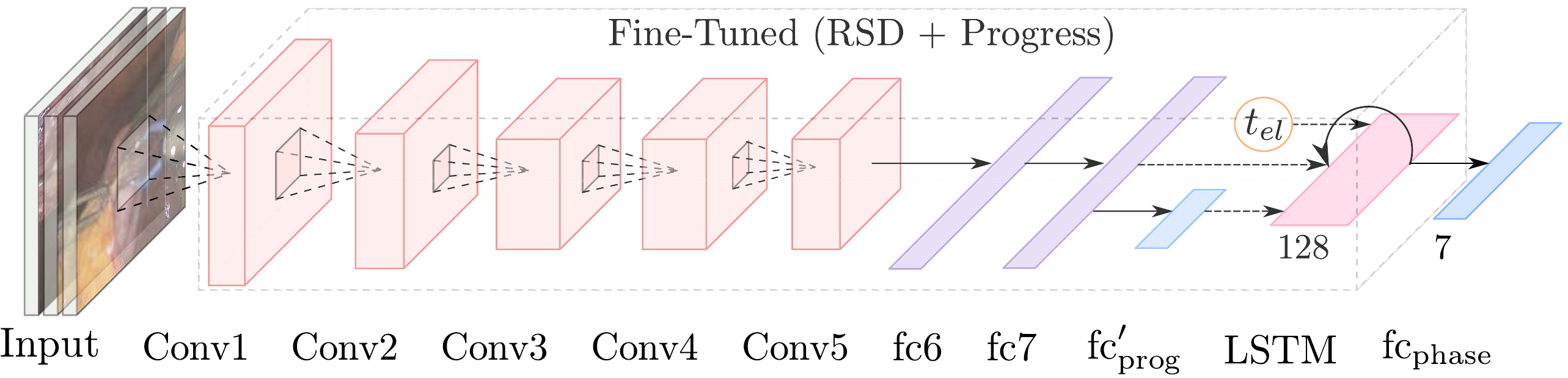}
        \caption{Updated \textit{EndoN2N} Model}\label{fig:endon2n_updated}  
        \end{subfigure}
        \end{center}
        \caption{Proposed RSD pre-training model. CNN architecture for progress regression is shown in (a). CNN-LSTM network designed for multi-task RSD and progress regression, incorporating elapsed time and predicted progress as additional LSTM features, is depicted in (b). (c) is the updated \textit{EndoN2N} CNN-LSTM network for compatibility with the RSD pre-training model. The layers within the dotted lines are fine-tuned after having been initialized on the training task mentioned within the brackets. The layers outside the dotted lines are randomly initialized.}\label{fig:RSD_pretraining}
    \end{figure*}
    
	\subsection{\textit{EndoN2N}}
	
	The \textit{EndoN2N} model, depicted in Figure \ref{fig:cnn_lstm}, combines a CNN with a LSTM. The model is adaptable to any CNN architecture and RNN variant. We utilize the LSTM as our recurrent network due to its robustness to the vanishing gradient problem \citep{hoc1997, bengio1994vanish}. Our experiments are performed using the  CaffeNet \citep{Jia2014caffe} CNN architecture, which is a slight modification of the AlexNet \citep{kri2012alexnet} architecture. Although we observed in our preliminary experiments that a deeper CNN architecture improves performance, we utilize CaffeNet, a relatively shallow architecture, for two reasons: (1) our aim is to provide an apples-to-apples comparison between the end-to-end training approach of the proposed \textit{EndoN2N} model and the two-step training approach of \textit{EndoLSTM}. This only requires a common CNN architecture to be used in both models and is not dependent on any specific architecture. Additionally, the advantages of the proposed RSD pre-training approach can also be demonstrated using any CNN architecture. (2) end-to-end CNN-LSTM training is computationally intensive. The use of a relatively shallow CNN makes it possible to perform an extensive experimental evaluation in order to clearly demonstrate the advantages of our proposed semi-supervised approach.
    
    
	
	As the CNN-LSTM network is trained in an end-to-end manner, we have named the model \textit{EndoN2N}. Before the end-to-end training step, the CNN is first separately fine-tuned for surgical phase recognition, depicted in Figure \ref{fig:cnn_finetuning}, in order to initialize the CNN-LSTM network with informative features corresponding to the surgical phases. In our experiments, we observe that the CNN-LSTM training converges to a poor local optima unless the CNN is first independently fine-tuned for surgical phase recognition. 
The CNN fine-tuning is accomplished by replacing the final fully connected layer of the CaffeNet architecture with a fully connected layer, $\textnormal{fc}'_{phase}$ as shown in Figure \ref{fig:cnn_finetuning}, of size equal to the number of surgical phases. In our case, there are $7$ output neurons. All the fine-tuned layers of the CNN, except $\textnormal{fc}'_{phase}$, are then appended to an LSTM which in turn is followed by a new fully connected layer, $\textnormal{fc}_{phase}$, also containing as many output neurons as the number of surgical phases (Figure \ref{fig:cnn_lstm}). The softmax function is applied at the end of both $\textnormal{fc}'_{phase}$ and $\textnormal{fc}_{phase}$ layers to obtain a probability distribution over the different surgical phases.
   
   Since surgical phase recognition is formulated as a multi-class classification problem,  we compute the classification loss using the multinomial logistic function defined as:
   \begin{equation}
   \mathcal{L} = \frac{-1}{T}\sum_{t=1}^{T}\sum_{p=1}^{M}y^{t}_{p}\log(\sigma(z^{t})_{p}),
   \label{eq:loss}
   \end{equation}
   where $T$ is the total number of frames in a laparoscopic video, $M$ refers to the number of distinct surgical phases, $y^{t}_{p}\in\{0,1\}$ is the ground truth for phase $p$ and $z^{t}$ is the vector of activations of $\textnormal{fc}_{phase}$ at the $t^{\textnormal{th}}$ time step of the surgery and $\sigma(\cdot)_{p}$ is the softmax function computing the predicted probability of phase $p$.
    
    \subsubsection{Training Approach}\label{sec:training}
    
    Here we present a detailed explanation of our end-to-end training approach for optimizing CNN-LSTM networks on long duration video sequences. The aim is to contrast our approach with the training approach employed in \textit{EndoLSTM} and existing approaches for end-to-end training of CNN-LSTM networks on laparoscopic video subsequences. Since the following discussion focuses on the approximation in the BPTT algorithm, a generic description of the layer-wise gradient computation is presented. While the discussion is based on the basic stochastic gradient descent algorithm for simplicity, any other optimization algorithm can also be used.
	    
    As illustrated in Figure \ref{fig:graph}, end-to-end training of a CNN-LSTM network requires the loss to be backpropagated through both the LSTM and CNN. Additionally, the BPTT algorithm requires the loss to be backpropagated from the last time step of a sequence, $t=T$, to the very first time step, $t=1$. If we denote the CNN weights as $W_{cnn}$ and the weights belonging to the LSTM as $W_{lstm}$, the gradients of the loss function $\mathcal{L}$, with respect to the network weights at a time instant $t$ can be expressed as:
        
    \begin{equation}\label{eq:lstm_grad}
    \frac{\partial\mathcal{L}}{\partial W_{lstm}}^{t} = f\left(P_t , \frac{\partial\mathcal{L}}{\partial W_{lstm}}^{t+1}\right),
    \end{equation}
    
    \begin{equation}\label{eq:cnn_grad}
    \frac{\partial\mathcal{L}}{\partial W_{cnn}}^{t} = g\left(\frac{\partial\mathcal{L}}{\partial W_{lstm}}^{t}\right),
    \end{equation}
    where $f$ and $g$ are generic functions used to express the computation of gradients for different layers of the LSTM and CNN and $P_{t}$ is the network prediction at the $t^{\textnormal{th}}$ time step as illustrated in Figure \ref{fig:graph}.  
    The boundary condition of the BPTT algorithm at the end of the sequence is:
    
     \begin{equation}\label{eq:bc_true}
    \frac{\partial\mathcal{L}}{\partial W_{lstm}}^{T+1} = 0.
    \end{equation}
    
    \begin{figure}[t]
    \begin{subfigure}[t]{0.48\textwidth}
    \centering
    \includegraphics[width=\textwidth]{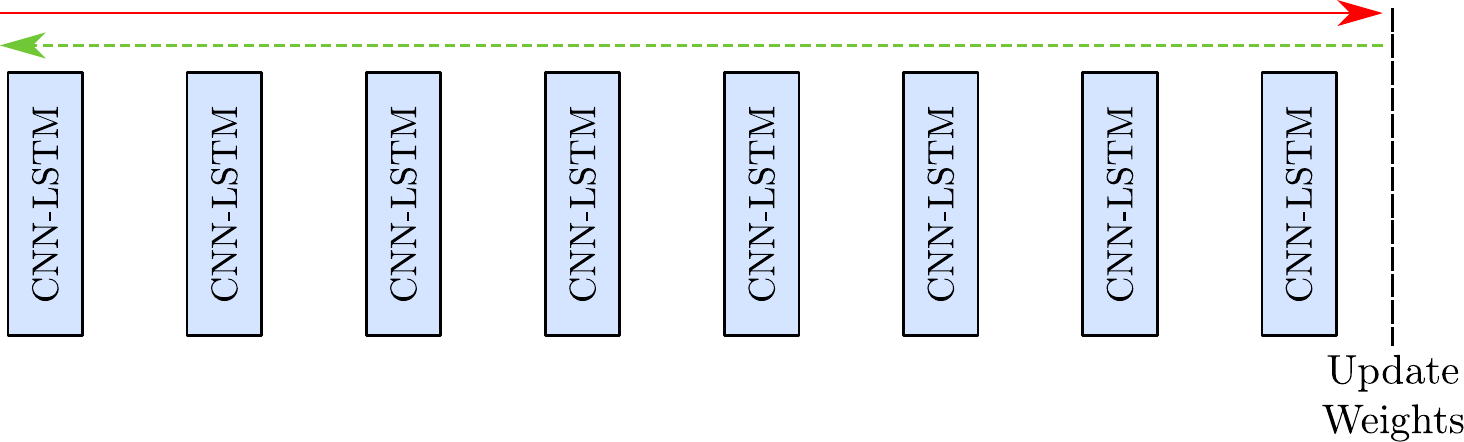}
    \caption{}\label{fig:true_backprop}
    \end{subfigure}
    \begin{subfigure}[t]{0.48\textwidth}
    \centering
    \vspace{0.25cm}
    \includegraphics[width=\textwidth]{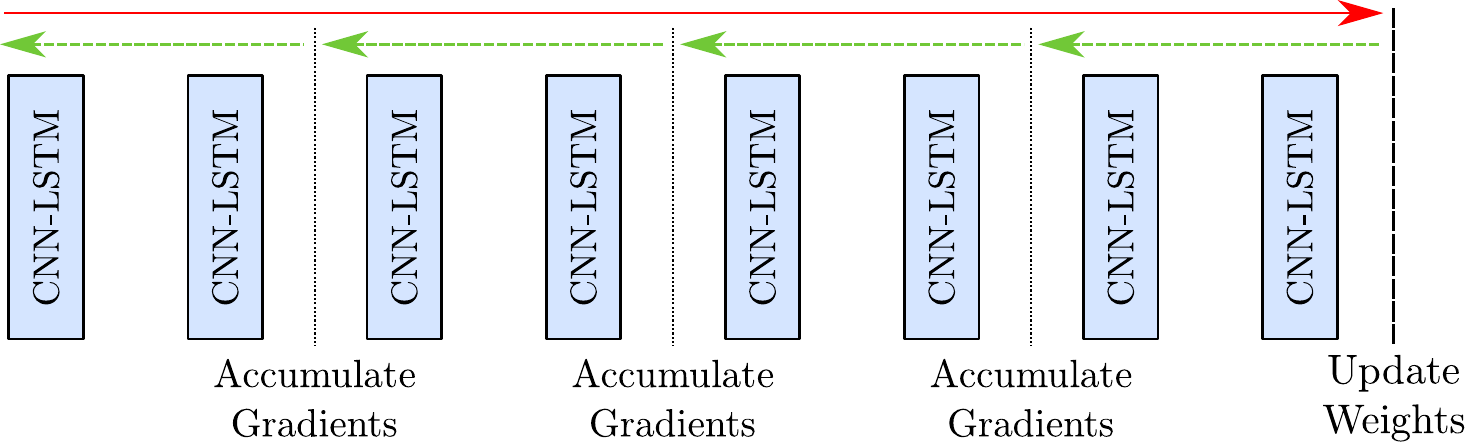}
    \caption{}\label{fig:backprop}
    \end{subfigure}
    \caption{Illustration of the BPTT algorithm. Red arrow denotes a forward pass and a green arrow denotes a backward pass. (a) depicts the standard algorithm and (b) illustrates our approximation.}
    \end{figure}
    
    The stochastic gradient descent weight update when utilizing a mini-batch of one video sequence of length $T$ is given by:
    \begin{equation}\label{eq:sgd_backprop}
    W^{\tau+1} = W^{\tau} - \eta\sum_{t=1}^{T}\frac{\partial\mathcal{L}}{\partial W^{\tau}}^{t},
    \end{equation}
    where $W^{\tau}$ are the learned CNN-LSTM weights at the end of $\tau$ training iterations. The weights are updated using the gradients computed for the entire video sequence, as shown in Figure \ref{fig:true_backprop}. The recursive structure of Equation \eqref{eq:lstm_grad} implies that to calculate the gradient of the loss function with respect to the network weights, $W$, at the first time step of the sequence, we require the gradients from the final time step. For this to be possible, the entire unrolled computational graph of the CNN-LSTM network, Figure \ref{fig:unrolled}, needs to be stored in memory. Due to the long duration of cholecystectomy surgeries and the large number of CNN parameters, end-to-end training of a CNN-LSTM network on complete laparoscopic video sequences has a high space complexity. Since no efficient method for storing the complete unrolled graph of the CNN-LSTM network during training is known, we utilize an approximation of the BPTT algorithm.
    
    In our approach, we restructure the loss shown in Equation \eqref{eq:loss} as:
    
     \begin{equation}
        	\mathcal{L} = \frac{-1}{\ell}\sum_{k=1}^{\ell}\frac{\ell}{T}\sum_{t=\frac{(k-1)T}{\ell}+1}^{\frac{kT}{\ell}}\sum_{p=1}^{M}y^{t}_{p}\log(\sigma(z^{t})_{p}).
            \label{eq:new_loss}
     \end{equation}    	
        
In Equation \eqref{eq:new_loss} we divide the complete laparoscopic video into $\ell$ consecutive subsequences. $\ell$ is appropriately selected such that the available computational resources are sufficient for storing the unrolled CNN-LSTM graph for $\sfrac{T}{\ell}$ time-steps. The loss is backpropagated for every subsequence and the gradients are accumulated independently over the $\ell$ different subsequences before updating the weights.

In this method, at the boundaries between consecutive subsequences, the LSTM cell states and hidden states are forward propagated, while the BPTT algorithm is truncated as illustrated in Figure \ref{fig:backprop}. This implies an approximation in Equation \eqref{eq:lstm_grad} at the subsequence boundaries as:

	\begin{equation}\label{eq:approx_lstm}
    \frac{\partial\mathcal{L}}{\partial W_{lstm}}^{\frac{kT}{\ell}} = f(P_{\frac{kT}{\ell}},0), \hspace{0.5cm}\forall k = 1,2,...,\ell \\
    \end{equation}

Since the gradient of the loss with respect to the weights of the CNN, Equation \eqref{eq:cnn_grad}, are dependent on the loss gradient with respect to the LSTM weights, these are being approximated as well. The stochastic gradient descent weight update step is now computed as:
	\begin{equation}\label{eq:sgd_backprop2}
    W^{\tau+1} = W^{\tau} - \eta\sum_{k=1}^{\ell}\sum_{t=\frac{(k-1)T}{\ell}+1}^{\frac{kT}{\ell}}\frac{\partial\mathcal{L}}{\partial W^{\tau}}^{t}.
    \end{equation}
    
        \subsection{\textit{EndoLSTM}}
	
\begin{figure}[t]
		\begin{subfigure}[ht]{0.15\textwidth}
    	\centering
		\includegraphics[height = 2.5cm]{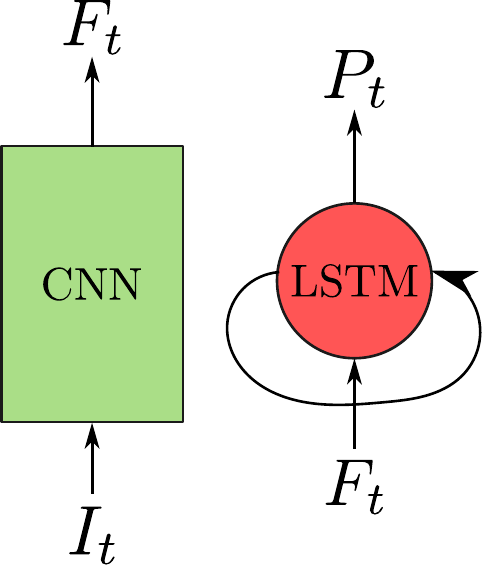}
    	\label{fig:lstm_rolled}
		\end{subfigure}
        \begin{subfigure}[ht]{0.35\textwidth}
    	\centering
		\includegraphics[height = 2cm]{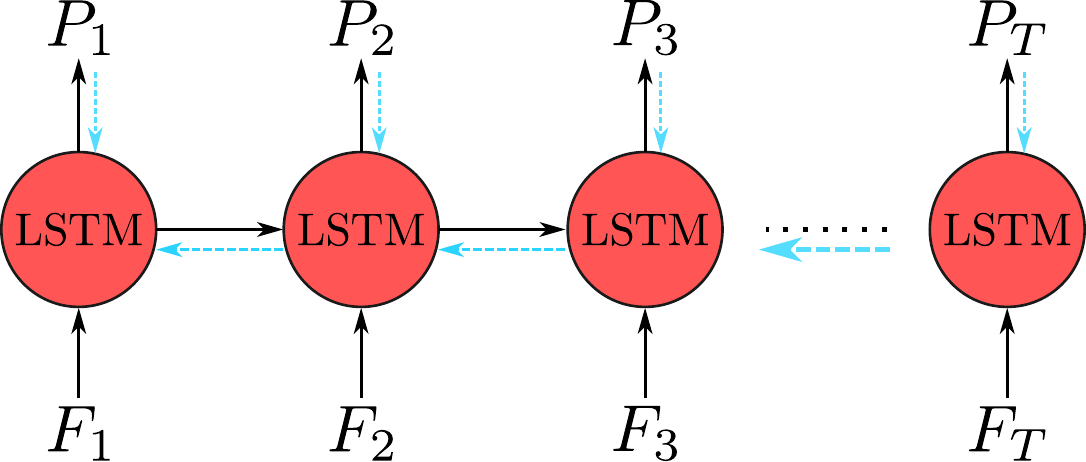}
    	\label{fig:lstm_unrolled}
		\end{subfigure}
        \caption{Computation graph for two-step optimization of a CNN-LSTM network. $I_t$ and $P_t$ are the input frame and network prediction at the $t^{\textnormal{th}}$ time-step of a sequence, respectively. $F_t$ is the extracted CNN features corresponding to $I_t$. The rolled graph is shown on the left. The unrolled LSTM network is depicted on the right.}\label{fig:lstm_graph}
    \end{figure}

	The architecture adopted for the \textit{EndoLSTM} model in our experiments is exactly the same as that of the \textit{EndoN2N} model, Figure \ref{fig:endon2n}. This is essential to be able to provide an accurate comparison between the two models. Similar to the \textit{EndoN2N} model, the CNN is first fine-tuned for surgical phase recognition to provide the LSTM with informative features. The difference between the two models lies in the backpropagation of the gradients through the CNN-LSTM network. In the \textit{EndoLSTM} model, only the weights of the LSTM are updated using the BPTT algorithm, while the weights of the CNN remain fixed, as depicted in Figure \ref{fig:lstm_graph}. Since the LSTM does not contain a large number of parameters like a CNN, it is feasible to store in memory the computational graph of the unrolled LSTM network over complete cholecystectomy video sequences. Hence, we do not need to approximate the BPTT algorithm.
    
	\subsection{Remaining Surgery Duration Pre-training}
	
    
	A key contribution of this work is the self-supervised pre-training of CNN-LSTM networks on the RSD prediction task. We hypothesize that accurate prediction of the time remaining in a surgery requires a good understanding of the surgical workflow. It is likely that a network that has been trained to accurately predict RSD would have indirectly gained knowledge related to the different surgical phases that occur, the duration of each phase and the variations in these surgical phases, since it would correspond to variations in remaining surgery duration. This could make it easier for the network to later be adapted for surgical phase recognition, thereby requiring less manually annotated data and making it easier to scale up surgical phase recognition to many types of surgeries.
	
	The RSD prediction task is formulated as a supervised regression task, where the network is provided with labels for the remaining surgery duration. Since for a given laparoscopic video the remaining surgery duration at a time instant is simply the remaining time in that video, the labels are available without the need for any manual annotation. As the labels are obtained for \textit{free}, RSD prediction is a self-supervised learning task. This makes it feasible to utilize a large number of laparoscopic videos to train a network for RSD prediction, ensuring that potentially valuable information from even unlabeled videos is exploited. For example, the network could acquire knowledge related to variable patient conditions and surgeon styles, thereby making it generalize better.
	
	\subsubsection{RSD Prediction Model}
	
    The CNN-LSTM network for RSD prediction is shown in Figure \ref{fig:rsdnet2}. The model is similar to the RSDNet model presented in \cite{twinanda2018rsd}, but for two key changes: (1) elapsed time and predicted progress are taken as additional input features into the LSTM and (2) the CNN-LSTM network is trained in an end-to-end manner. Although the original RSDNet model uses a two-step optimization, similar to the \textit{EndoLSTM} model, end-to-end training is the most natural choice for pre-training a CNN-LSTM network. End-to-end training enables the optimal correlation between the features learned by the CNN and by the LSTM.
	
	We adopt the approach proposed by \cite{twinanda2018rsd} to learn a RSD prediction model without any manual annotation, which is contrary to previous approaches \citep{aksamentov2017miccai}. This involves first the fine-tuning of the CNN for progress estimation, as depicted in Figure \ref{fig:progress_cnn}, which is the task of predicting the percentage of the surgery that has been completed at a given time instant. Progress estimation is also formulated as a self-supervised regression task. The CNN-LSTM model is then trained for the multi-task objective of RSD prediction and progress prediction. \cite{twinanda2018rsd} showed that training for this multi-task objective was better than training for RSD alone.
    
	End-to-end training of a CNN-LSTM network on laparoscopic video sequences for RSD prediction is performed with the same approach used in \textit{EndoN2N} (Section \ref{sec:training}). We restructure the loss function as:
    \begin{equation}
    	\begin{split}
        	\mathcal{L} &= \frac{-1}{T}\sum_{t=1}^{T}\left[y^{t}_{rsd}\Omega(z^{t}_{rsd}) + y^{t}_{prog}\Omega(\rho(z^{t}_{prog}))\right],\\
            &= \frac{-1}{\ell}\sum_{k=1}^{\ell}\frac{\ell}{T}\sum_{t=\frac{(k-1)T}{\ell}+1}^{\frac{kT}{\ell}}\left[y^{t}_{rsd}\Omega(z^{t}_{rsd}) + y^{t}_{prog}\Omega(\rho(z^{t}_{prog}))\right],
            \label{eq:smooth_l1_loss}
        \end{split}
    \end{equation}
    where $y^{i}_{rsd}$ and $y^{i}_{prog}$ are the ground truths for RSD and progress, $z^{i}_{rsd}$ and $z^{i}_{prog}$ are the activations of the fully connected layers $\textnormal{fc}_{\textnormal{rsd}}$ and $\textnormal{fc}_{\textnormal{prog}}$ for the $t^{\textnormal{th}}$ frame of the laparoscopic video. $\rho(\cdot)$ is the sigmoid function, and $\Omega$ is the smooth L1 loss \citep{girshick2015} defined as:
    
    \begin{equation}
    	\Omega(x) = \begin{dcases}
    		0.5x^{2},& \textnormal{if } |x|<1 \\
            |x|-0.5,& \textnormal{otherwise}.
    	\end{dcases}
    \end{equation}
	
	\subsubsection{Updated Surgical Phase Recognition Model}
	
    In \cite{twinanda2018rsd}, it was argued that knowledge of the elapsed time ($t_{el}$) and progress ($prog$) was beneficial for RSD prediction, since they possess a fundamental relation with RSD ($t_{rsd}$), as shown below:
	\begin{equation}\label{eq:time}
		t_{rsd} = T - t_{el} = \frac{t_{el}}{prog} - t_{el},
	\end{equation}
	where $T$ is the total duration of the surgery. The RSDNet model simply concatenated elapsed time along with the output of the LSTM and incorporated progress only as an additional output to be predicted. We incorporate elapsed time as well as estimated progress, $\textnormal{fc}'_{\textnormal{prog}}$, as input features to the LSTM itself as shown in Figure \ref{fig:rsdnet2}. We believe the LSTM is then capable of learning more complex relationships in between the elapsed time and the model's perception of surgery progress and RSD.
    
	The goal of utilizing these additional features in our CNN-LSTM model is to make RSD prediction a more effective pre-training approach for surgical phase recognition. The \textit{EndoN2N} model, however, needs to be modified in order to be compatible with the proposed RSD pre-training model. The updated \textit{EndoN2N} model architecture, which includes these additional features as inputs to the LSTM, is shown in Figure \ref{fig:endon2n_updated}. We later present an ablation study to demonstrate the advantages of proposed RSD pre-training model and the updated \textit{EndoN2N} architecture.
	
	\subsection{Temporal Context Pre-training}
	
    \begin{figure}[t]
		\centering
        \includegraphics[width=4.5cm]{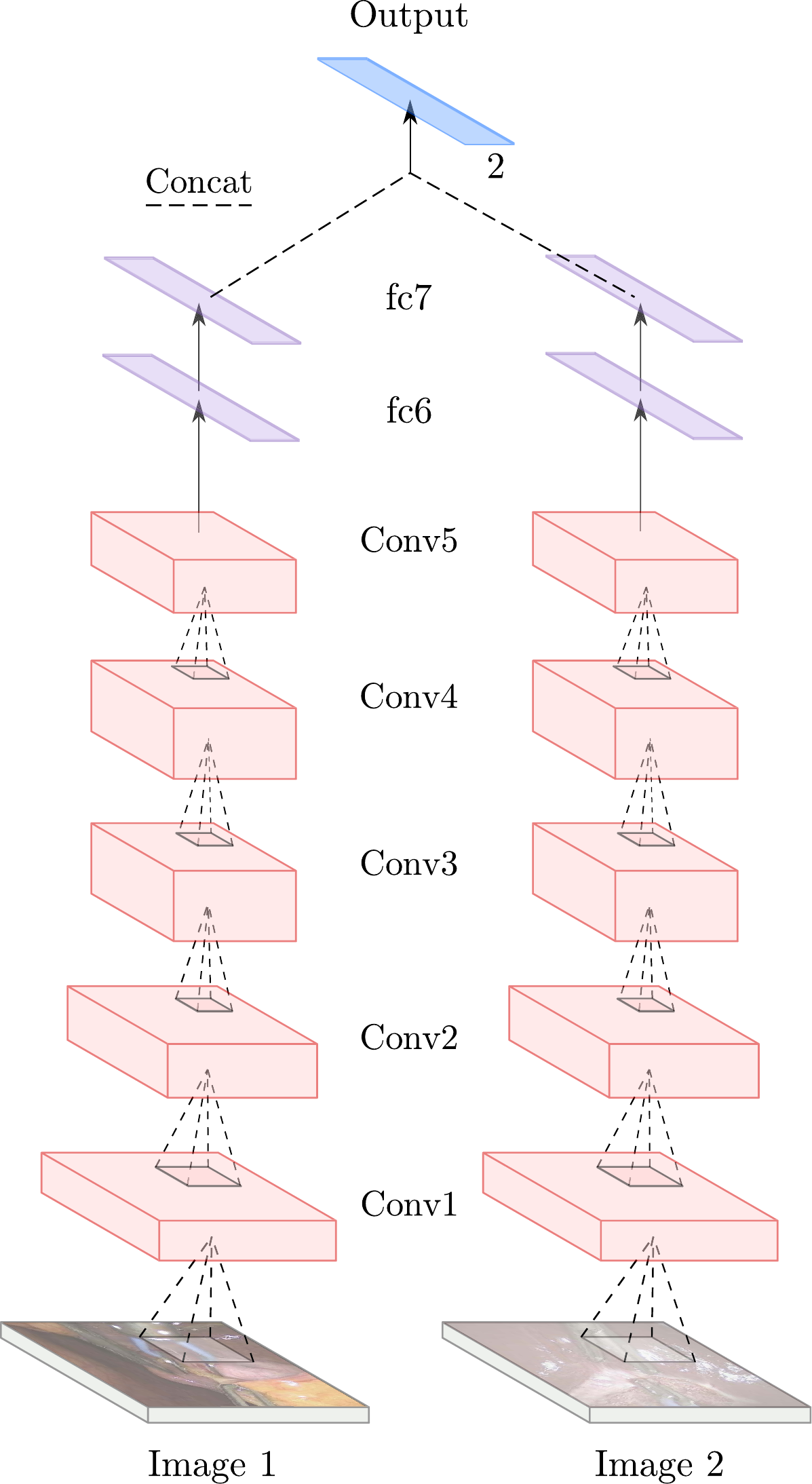}
        \caption{TempCon pre-training model.}\label{fig:tempcon}    
    \end{figure}
    
    The TempCon pre-training approach of \cite{bod2017}, which we use as a baseline for comparison, aims to learn the temporal order of laparoscopic workflow by training a Siamese network to predict the relative order of two randomly sampled frames of a laparoscopic video. The specific model used in our experiments is derived from the architecture of \cite{misra2016unsupervised}, which is designed for a similar task of predicting the correct temporal order of randomly sample frames, since we observed it to perform better than the architecture of \cite{bod2017} in our preliminary experiments ($74$ percent vs $72$ percent accuracy on the temporal context prediction task). Additionally, such an architecture ensures that all the layers of the CaffeNet CNN used in the subsequent surgical phase recognition task will be pre-trained.
	
	This approach is designed for pre-training CNNs only. A two-stream Siamese network is created by replicating layers conv1 to fc7 of the CaffeNet architecture, as shown in Figure \ref{fig:tempcon}. Two randomly sampled frames from a laparoscopic video are provided as inputs to the network. Weights are shared in between the two-streams. The final layers of the Siamese network are concatenated and are followed by a fully connected layer comprising of two neurons which provides the classification output. Each neuron respectively corresponds to one of the input frames. The output of the network is either 0 or 1 depending on whether frame 1 or frame 2 is predicted as occurring first in the surgical sequence.
       
       \begin{table*}[t]
\begin{centering}
\begin{tabular}{|l|c|c||c|c|c|c|c|c|c|}
\hline 
\multicolumn{2}{|c|}{\multirow{2}{*}{Model}} & \multirow{2}{*}{Task} &  \multicolumn{7}{c|}{Hyperparameters}\tabularnewline
\cline{4-10}
\multicolumn{2}{|c|}{} & & Optimizer & Iterations & $\alpha$ & Step-Size& $\gamma$ & Batch-Size & $\lambda$ \tabularnewline
\hline
\hline 
\multirow{2}{*}{\textit{EndoN2N}}  & CNN Fine-tuning & Phase & SGD & 50k & $10^{-3}$ & 20k & 0.1  & 50& 5$\cdot10^{-4}$\tabularnewline
\cline{2-10} 
 & CNN-LSTM training & Phase  & Adam & 8k & $10^{-4}$ & 2k & 0.25& 500 $\times$12 & 5$\cdot10^{-4}$\tabularnewline
\hline
\multirow{2}{*}{\textit{EndoLSTM}}  & CNN fine-tuning & Phase & SGD & 50k &$10^{-3}$ &20k & 0.1  & 50& 5$\cdot10^{-4}$\tabularnewline
\cline{2-10} 
 & LSTM training & Phase & SGD & 30k & $10^{-3}$& 10k& 0.1& 6000& 5$\cdot10^{-4}$\tabularnewline
\hline 
\hline
\multirow{2}{*}{RSD Prediction}  & CNN fine-tuning & Progress &  SGD & 50k &$10^{-3}$ &15k & 0.1 & 64 & 5$\cdot10^{-4}$\tabularnewline
\cline{2-10} 
 & CNN-LSTM training & RSD - Progress & SGD & 8k &$10^{-3}$ & 2k& 0.5& 500 $\times$12& $10^{-3}$\tabularnewline
\hline
\multicolumn{2}{|c|}{Temporal Context prediction} & Relative Order  & SGD & 50k & 5$\cdot10^{-4}$& n/a&n/a &160 &5$\cdot10^{-4}$ \tabularnewline
\hline
\end{tabular}\tabularnewline
\par\end{centering}
\vspace{0.2cm}
\caption{Training hyperparameters for each individual model, including their different training steps and their respective training task.\label{tab:hyperparameters}}
\end{table*}

	\section{Experimental Setup}
	    
	The experiments are carried out on $Cholec120$, a dataset of 120 cholecystectomy laparoscopic videos. The surgical procedures contained in the dataset were performed by 33 surgeons at the University Hospital of Strasbourg. The videos are recorded at 25 fps and have an average duration of 38.1 mins ($\pm$16.0 mins). In total, the dataset accumulates  over  75  hours  of recordings. All 120 videos have been annotated at a frame rate of 1 fps with surgical phase labels corresponding to the 7 phases shown in Figure \ref{fig:phases}.
	

	We have designed experiments for two specific goals: (1) to evaluate the improvement in performance obtained by the \textit{EndoN2N} model over \textit{EndoLSTM} and (2) to demonstrate the benefits of the proposed self-supervised RSD pre-training approach in reducing the reliance of supervised surgical phase recognition algorithms on annotated data. The division of data for the experiments and the evaluation metrics are described below.
	
	\subsection{\textit{EndoN2N} Evaluation}
	
	Surgical phase recognition performance is evaluated on the \textit{Cholec120} dataset using a 4-fold cross-validation setup. Each fold is divided into 80 training, 10 validation and 30 test videos. 60 randomly sampled training videos (75 percent) are used to first fine-tune the CNN individually for surgical phase recognition. All 80 videos are then used to train the combined CNN-LSTM network of the \textit{EndoN2N} model. In the case of the \textit{EndoLSTM} model, the LSTM is independently trained on the 80 training videos by utilizing features extracted from the fine-tuned CNN. All 10 validation videos are used during CNN fine-tuning as well as CNN-LSTM or just LSTM training in case of \textit{EndoN2N} or \textit{EndoLSTM}, respectively. The final \textit{EndoN2N} model weights selected for testing correspond to the best performing model on the validation set. The validation videos are also used to perform the hyperparameter search discussed in Section \ref{sec:model_training}. Both models are evaluated on all 30 of the test videos. The final results presented are the averages over the four folds.
	
	\subsection{RSD Pre-training Evaluation}\label{sec:pretraining_data}
	
	The \textit{EndoN2N} model is utilized to evaluate the advantages of the proposed RSD pre-training approach in reducing the amount of annotated data required for successful surgical phase recognition. We perform a comparison between the following three pre-training approaches: (1) RSD pre-training, (2) TempCon pre-training and (3) no self-supervised pre-training. The experiments are conducted using the same 4-fold cross-validation setup. For each fold, all 80 training videos are used for pre-training the network, without relying on the available annotations. The \textit{EndoN2N} model is then fine-tuned for surgical phase recognition using 10, 20, 25, 40, 50, 80 and 100 percent of the labeled training videos from each fold, i.e., 8, 16, 20, 32, 40, 64 and 80 videos respectively. The 80 training videos of each fold are divided into four quartiles based on the surgery durations and the supervised fine-tuning subsets are created by sampling an equal number of videos from each quartile. 4 different subsets of 8, 16, 20, 32 and 40 videos along with 2 different subsets of 64 videos are sampled from each fold. The average performance over these different subsets is evaluated in order to ensure that the model is not biased by the particular videos selected. 75 percent of the total fine-tuning videos are randomly sampled in each case for the initial CNN fine-tuning step. The evaluation is again performed on the 30 test videos of each fold, and the final results are the averages over all four folds. 


	\subsection{Evaluation Metrics}
    
    To provide a quantitative measure of the performance of the proposed surgical phase recognition models, we utilize the metrics of accuracy, precision and recall as defined in \cite{padoy2012}. Accuracy is defined as the percentage of correct surgical phase predictions within a laparoscopic video. Precision is defined as the ratio between correct predictions and the total number of predictions, while recall is the ratio between correct predictions and the total number of instances in the ground truth. In every laparoscopic video, precision and recall are computed for each individual phase and the average values are reported as well. 
    
    To compare the performance of the various models utilizing self-supervised pre-training, we use the F1-score metric, which is the harmonic mean of the average precision and recall values, since it provides a balanced measure of the combined precision and recall metrics. The use of a single score allows us to concisely quantify the performance of different models and eases comparisons.
    
    \begin{table*}[t]
		\begin{center}
			\begin{tabular}{|c c|c|c|c|c|}
				\hline
				\multicolumn{2}{|c|}{\multirow{2}{*}{  Model  }} & \multirow{2}{*}{ Accuracy } & \multirow{2}{*}{ Average Precision } & \multirow{2}{*}{ Average Recall } & \multirow{2}{*}{ F1-Score } \tabularnewline
				\multicolumn{2}{|c|}{} & & & &\tabularnewline
				\hline
				\multicolumn{2}{|c|}{ \textit{EndoN2N} } & \textbf{86.7$\pm$9.3} & \textbf{81.4$\pm$23.0} & \textbf{80.9$\pm$22.1} & \textbf{81.1$\pm$7.5}\tabularnewline
				\hline
				\multicolumn{2}{|c|}{ \textit{EndoLSTM} } & 83.0$\pm$10.8 & 77.5$\pm$24.0 & 77.2$\pm$24.2 & 77.3$\pm$8.0\tabularnewline\hline
			\end{tabular}
		\end{center}
		\caption{Surgical phase recognition performance in terms of accuracy, average precision, average recall and F1-score (percentages) evaluated on the complete Cholec120 dataset. Results have been calculated using 4-fold cross validation.}\label{tab:full_cholec120}
	\end{table*}
    
    	\begin{table*}[t]
		\begin{subtable}{\linewidth}
			\begin{center}
			\begin{tabular}{|l| *{7}{c|}}
				\hline
				Precision & P1 & P2 & P3 & P4 & P5 & P6 & P7 \tabularnewline
				\hline
				\textit{EndoN2N} & \textbf{84.5$\pm$25.0} & \textbf{91.5$\pm$11.2} & \textbf{76.3$\pm$22.9} & \textbf{90.2$\pm$13.9} & \textbf{79.4$\pm$18.6} & 72.3$\pm$32.9 & \textbf{75.4$\pm$28.1} \tabularnewline
				\hline
				\textit{EndoLSTM} & 77.5$\pm$28.3 & 90.8$\pm$11.5 & 64.4$\pm$28.1 & 83.8$\pm$18.7 & 76.6$\pm$20.2 & \textbf{74.9$\pm$29.1} & 74.5$\pm$26.6 \tabularnewline
				\hline
			\end{tabular}
			\end{center}
            \vspace{0.2cm}
		\end{subtable}
		\begin{subtable}{\linewidth}
			\begin{center}
			\begin{tabular}{|l| *{7}{c|}}
				\hline
				Recall & P1 & P2 & P3 & P4 & P5 & P6 & P7 \tabularnewline
				\hline
				\textit{EndoN2N} & \textbf{84.5$\pm$24.0} & \textbf{93.7$\pm$11.4} & \textbf{70.6$\pm$26.0} & 90.2$\pm$14.3 & 77.8$\pm$19.1 & 71.1$\pm$29.3 & \textbf{78.8$\pm$24.8} \tabularnewline
				\hline
				\textit{EndoLSTM} & 71.4$\pm$28.5 & 87.8$\pm$18.6 & 63.2$\pm$29.6 & \textbf{90.6$\pm$18.1} & \textbf{78.2$\pm$18.4} & \textbf{73.6$\pm$28.5}  & 75.6$\pm$24.0 \tabularnewline
				\hline
			\end{tabular}
		\end{center}
		\end{subtable}
		\caption{Surgical phase recognition performance for each individual phase in terms of precision and recall metrics, evaluated on the complete Cholec120 dataset using 4-fold cross validation.}\label{tab:per_phase}
	\end{table*}
    
    \begin{figure*}[t]
	\begin{center}
	\begin{subfigure}[ht]{0.45\textwidth}
    \centering
	\includegraphics[height = 6cm]{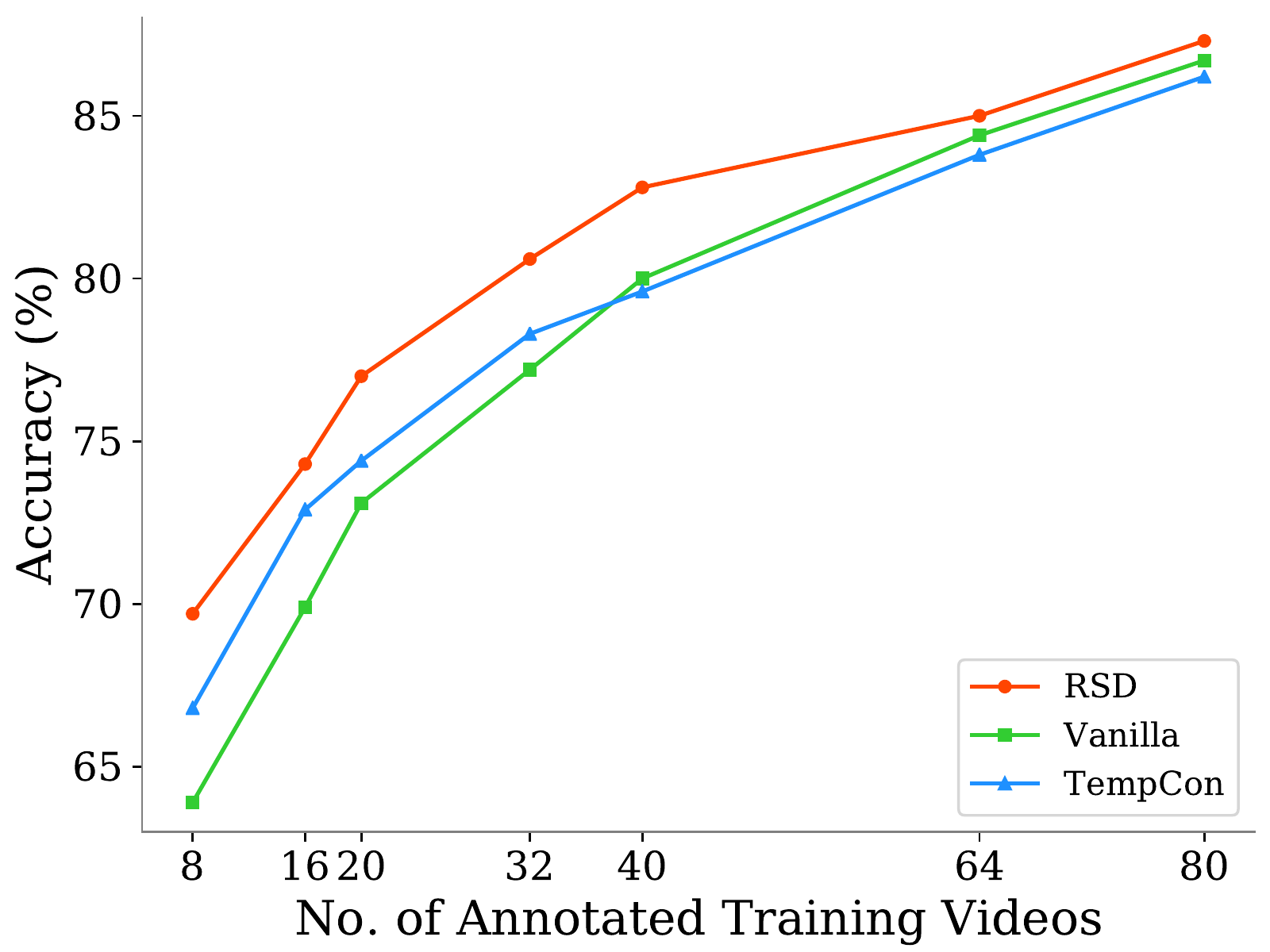}
    \caption{Accuracy}
	\end{subfigure}
    \begin{subfigure}[ht]{0.45\textwidth}
    \centering
	\includegraphics[height = 6cm]{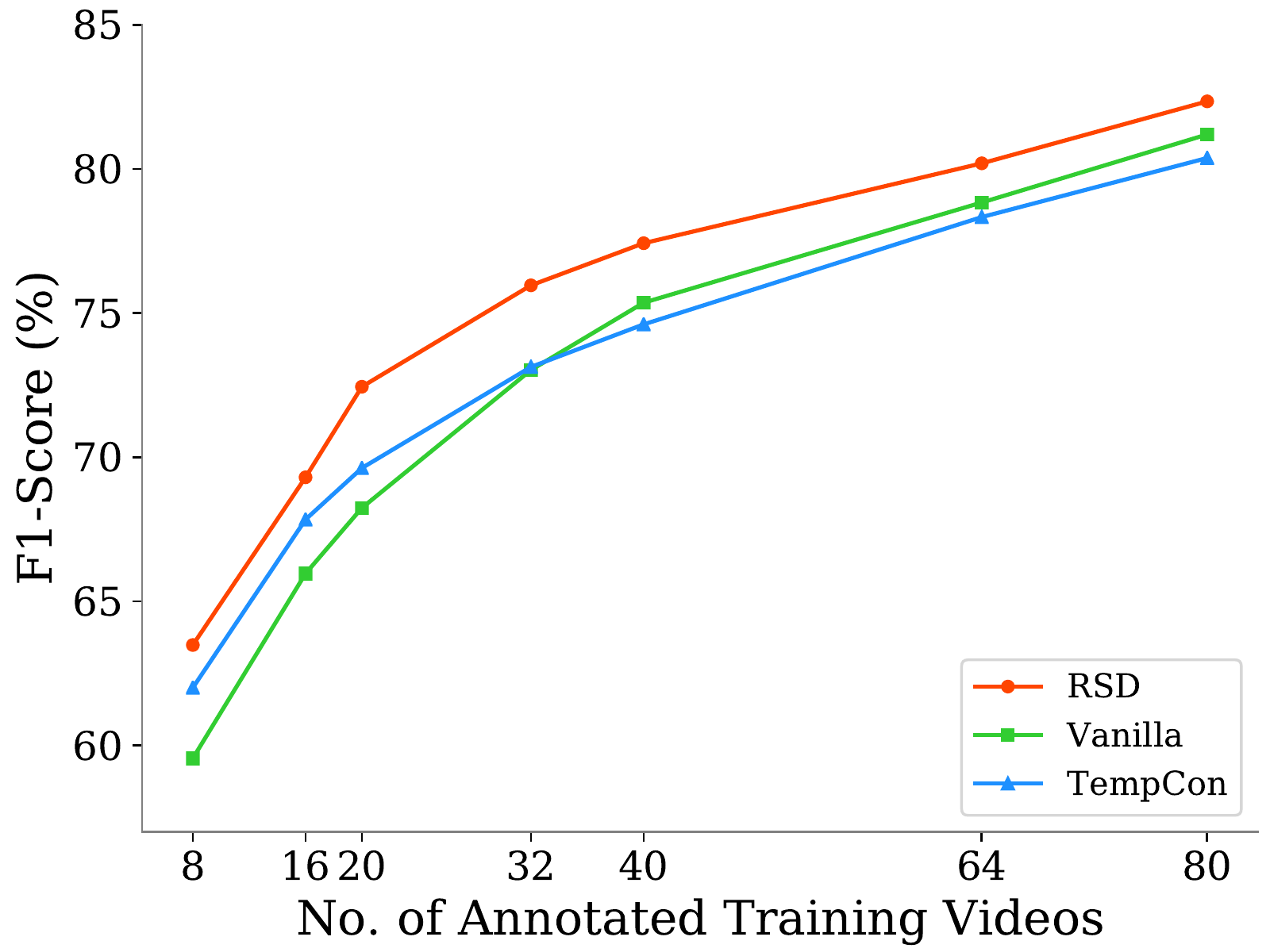}
	\caption{F1-Score}
	\end{subfigure}
	\par\end{center}
	\caption{Comparison of surgical phase recognition performance of the \textit{EndoN2N} model initialized using (1) only ImageNet pre-training without any self-supervised pre-training (vanilla \textit{EndoN2N}), (2) the proposed RSD pre-training or (3) temporal context pre-training. The effect of variation in the number of annotated training videos on surgical phase recognition performance in terms of (a) accuracy and (b) F1-score is illustrated.\label{fig:pretraining}}
	\end{figure*}
    
\section{Model Training}\label{sec:model_training}
	
All the experiments are performed using the \textit{Caffe} library \citep{Jia2014caffe}. In order to obtain an effective training setup for the \textit{EndoN2N} and RSD pre-training models, a hyperparameter search was performed over the optimizer as well as several parameter values such as the learning rate ($\alpha$), size of LSTM hidden state vectors, learning rate decay factor ($\gamma$), learning rate decay step-size and regularization factor ($\lambda$). Table \ref{tab:hyperparameters} details the hyperparameters for the different models discussed in this paper. It is to be noted that all layers with random initializations are set a 10 times higher learning rate than pre-trained layers and that weights of the $\textnormal{fc}'_{\textnormal{prog}}$ layer of the RSD-Progress multi-task network (Figure \ref{fig:rsdnet2}) are not updated. 

The stochastic gradient descent (SGD) optimizer is utilized with a momentum of 0.9 and the Adam optimizer is implemented with the parameters proposed in \cite{kingma2015adam}. While utilizing the Adam optimizer was beneficial for the \textit{EndoN2N} model, we did not find it effective for \textit{EndoLSTM}.
        
        
	\subsection{\textit{EndoN2N} Weight Initialization}\label{sec:initialization}
	
	In the experiment comparing the \textit{EndoN2N} and the \textit{EndoLSTM} models, the network weights are initialized from the open-source CaffeNet model, which has been pre-trained on the ImageNet dataset. No self-supervised pre-training is utilized. This initializes the CNN layers with pre-trained weights, while the LSTM is randomly initialized. We refer to this as the vanilla \textit{EndoN2N} model.
	
	For the self-supervised pre-training experiments, the \textit{EndoN2N} model is pre-trained on either the RSD prediction or TempCon prediction task. Transferring weights from a model trained for RSD prediction enables both the CNN and LSTM to be initialized with pre-trained weights. However, in the case of TempCon pre-training, only the CNN weights are pre-trained while the LSTM weights are once again randomly initialized. 

	
	\subsection{End-to-End CNN-LSTM Training}
	The batch size used in a single forward pass corresponds to subsequences of 500 consecutive frames. We trained our models on NVidia GeForce GTX TitanX and NVidia GeForce GTX 1080 GPUs, with 12 GB and 11GB of RAM respectively, which is sufficient for storing the complete unrolled CNN-LSTM graph for 500 time-steps. Since the longest video comprises of 5987 frames when sampled at 1 fps, all videos are padded with blank images to make them equal to 6000 frames. During training the loss is accumulated for 12 forward passes before performing a weight update. The padded images are ignored from the loss computation. Hence, one complete iteration corresponds to an effective batch size of one video. Additionally, the total iterations during end-to-end CNN-LSTM training always correspond to 100 epochs. The iterations and step-size are scaled proportionally when different amounts of videos are used for training as discussed in Section \ref{sec:pretraining_data}.
    
	

	\subsection{RSD Pre-training}
	
	The CNN is first trained for progress estimation after initializing the weights from the CaffeNet model pre-trained on the ImageNet dataset. Unlike in the phase recognition pipeline, data from all training videos are used for fine-tuning the CNN with self-supervision since this leads to optimal semi-supervised surgical phase recognition performance.
	
	As discussed in \cite{twinanda2018rsd}, the naturally high range of RSD target values for cholecystectomy surgeries (longest surgery in \textit{Cholec120} being 100 minutes) would need to be normalized in order to be able to regress the target values, while using a sufficiently large regularization parameter to prevent overfitting. We use the same normalization factor that was used by \cite{twinanda2018rsd} on the Cholec120 dataset, \ie $s_{norm}=5$.

	\subsection{TempCon Pre-training}
	
	The CNN weights are first initialized on the ImageNet dataset similar to the RSD pre-training network. Unlike our proposed RSD pre-training, which utilizes complete video sequences, TempCon pre-training requires pairs of frames to be sampled from the videos. From each of the 80 training videos, 50k pairs of frames are sampled. The model is trained for two full epochs over all the sampled pairs of frames.
	
\section{Results}
	
	\subsection{\textit{EndoN2N} Evaluation}
	
    Table \ref{tab:full_cholec120} shows a comparison between the accuracy as well as average precision and recall across all surgical phases obtained by the \textit{EndoN2N} and \textit{EndoLSTM} models on the complete Cholec120 dataset with the 4-fold cross-validation setup. \textit{EndoN2N} outperforms \textit{EndoLSTM} in each of the metrics.
	
	A comparison between the recognition performance of \textit{EndoN2N} and \textit{EndoLSTM} for each of the individual surgical phases is shown in Table \ref{tab:per_phase}. As expected from the results of Table \ref{tab:full_cholec120}, \textit{EndoN2N} performs better in most of the phases in terms of both precision and recall. The 3rd phase, the clipping and cutting phase, is the most crucial phase of cholecystectomy surgeries. It is also a short duration phase and occurs in between two of the longest duration phases, making it difficult for a surgical phase recognition algorithm to recognize. \textit{EndoN2N} is seen to be considerably better at recognizing this phase. It can also be seen that even in the few cases that \textit{EndoLSTM} outperforms \textit{EndoN2N}, the difference is not significant in any of the metrics.
	
	\subsection{RSD Pre-training Evaluation}
	    
    The graphs depicted in Figure \ref{fig:pretraining} illustrate the variation in surgical phase recognition performance with different amounts of annotated training data. The proposed RSD pre-training approach (shown in red) leads to superior performance for all sets of training data in terms of both accuracy and F1-score. TempCon pre-training is only effective when the ratio of the quantity of annotated training videos to pre-training videos is small. When the number of annotated videos increases, the pre-training approach starts to become detrimental, which is a common trend in semi-supervised learning \citep{paine2014}. On the other hand, the proposed RSD pre-training improves performance even when the training data is fully annotated, further highlighting the superiority of the approach.
    
            \begin{figure}[t]
		\begin{subfigure}[ht]{0.5\textwidth}
    	\centering
		\includegraphics[width = 7cm]{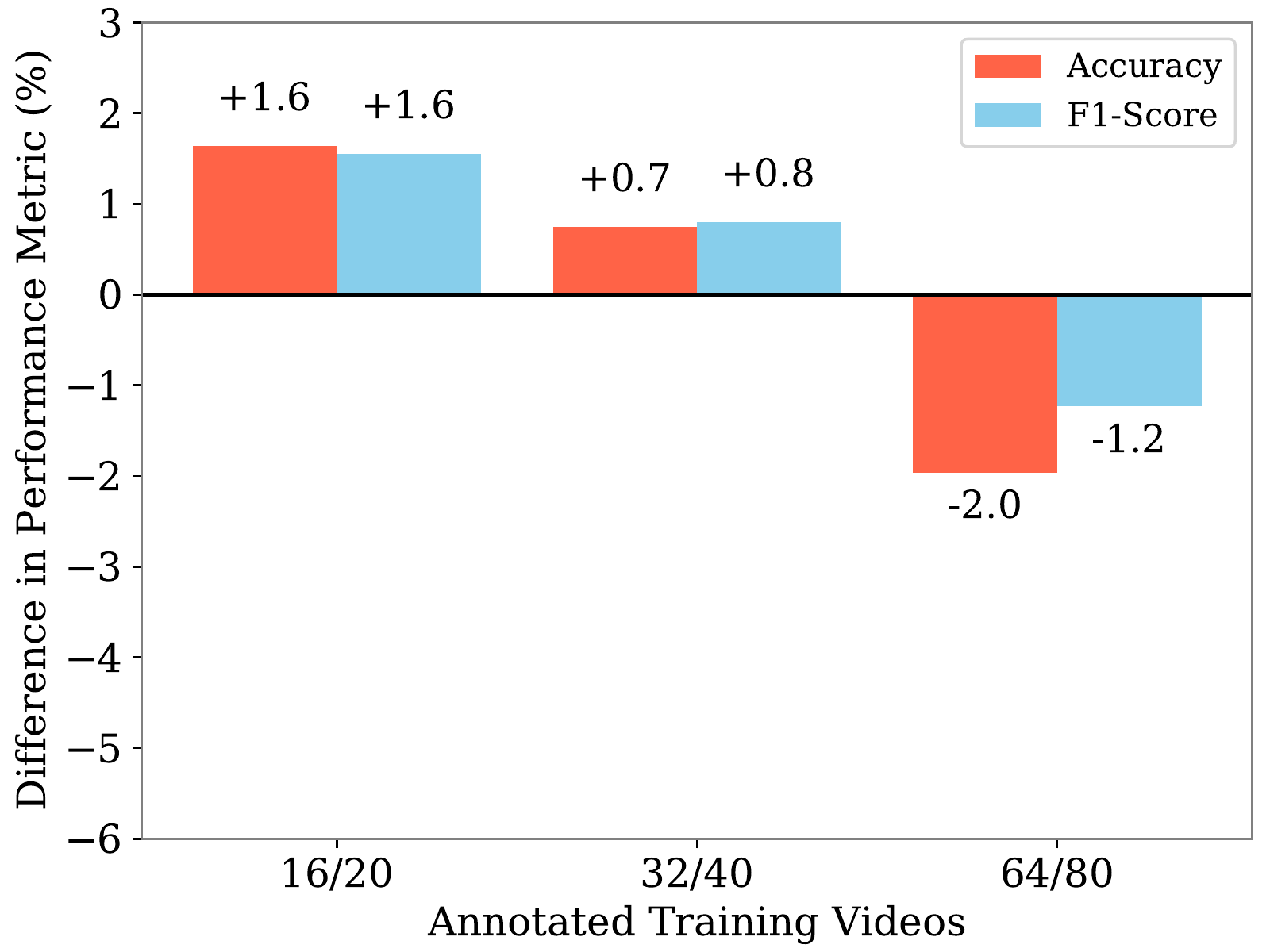}
    	\caption{}\label{fig:80ratio}
		\end{subfigure}
        \begin{subfigure}[ht]{0.5\textwidth}
    	\centering
		\includegraphics[width = 7cm]{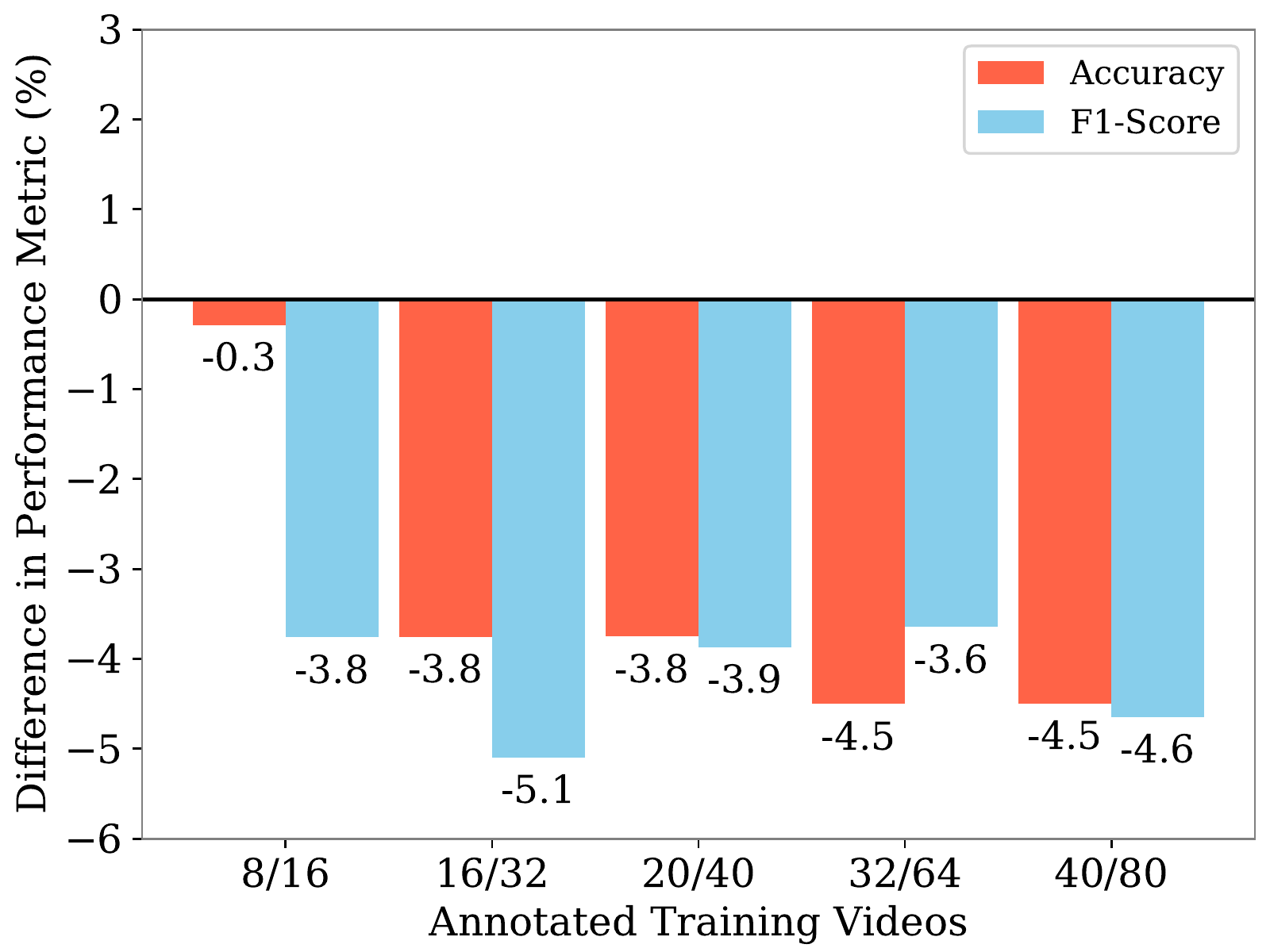}
    	\caption{}\label{fig:50ratio}
		\end{subfigure}
        \caption{Relative performance of the RSD pre-trained \textit{EndoN2N} model with respect to the vanilla \textit{EndoN2N} model, derived from Figure \ref{fig:pretraining}. The RSD pre-trained model is supervised using either (a) 20\% or (b) 50\% fewer annotated videos than the vanilla \textit{EndoN2N} model. The pair of numbers on the horizontal axis represents the number of annotated training videos used by the RSD pre-trained model/vanilla \textit{EndoN2N} model, respectively.}\label{fig:ratios}
    \end{figure}
    
	To highlight the effectiveness of the proposed RSD pre-training approach in reducing the reliance of surgical phase recognition models on annotated laparoscopic videos, we show in Figure \ref{fig:ratios} the relative performance of the RSD pre-trained \textit{EndoN2N} model, trained using less annotated videos, as compared to the same model without any self-supervised pre-training, but trained on more annotated videos. We notice that similar levels of performance can be achieved with less annotated data by adopting our proposed pre-training approach. Figures \ref{fig:80ratio} and \ref{fig:50ratio}, which are derived from Figure \ref{fig:pretraining}, show the difference in performance when (a) 20\% and (b) 50\% fewer annotated training videos are utilized respectively. The pre-training is still performed using 80 videos. In Figure \ref{fig:80ratio}, the RSD pre-trained model using less labeled videos performs better in general when the number of pre-training videos is higher than the number of annotated videos. This is of particular significance for actual clinical application, where there is a vast amount of data, but only a small fraction of it can be annotated. In Figure \ref{fig:50ratio} we see that the accuracy drops further as the number of labeled videos increases. This is expected since pre-training is more effective when the ratio of the amount of pre-training data to annotated data is high. Yet, the difference in accuracy is still always under 5\%, even though only half the number of annotated videos are used after the pre-training. The difference in F1-score also remains within a similar range, where the largest difference observed is 5.1\%.

	\section{Discussion}
	
	

    
	\subsection{Ablation Study}
    
     \begin{figure}[ht]
    	\centering
        \includegraphics[width=7cm]{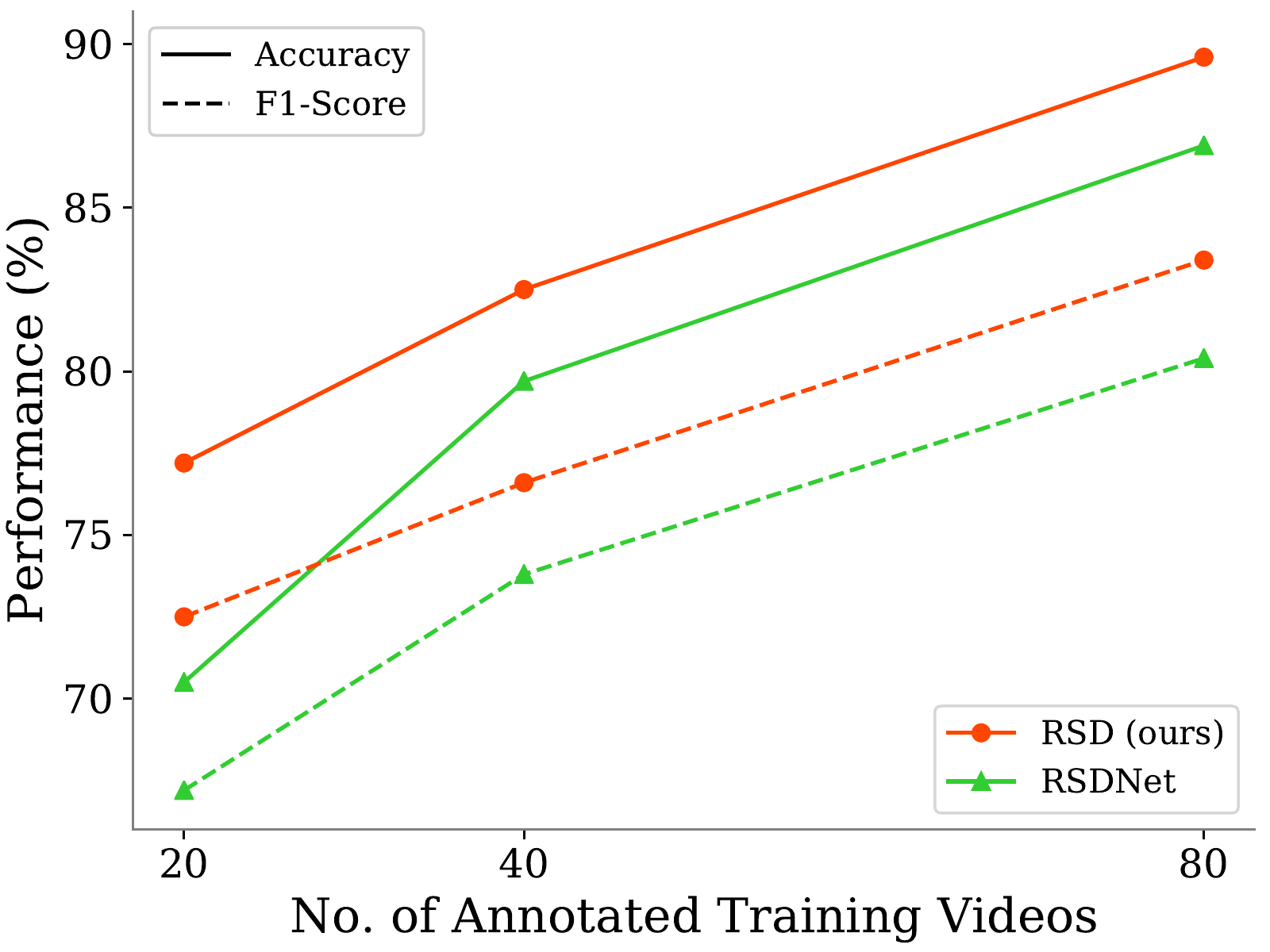}
        \caption{Comparison between the surgical phase recognition performance of \textit{EndoN2N} when pre-trained using either our proposed RSD prediction architecture or the RSDNet architecture of \cite{twinanda2018rsd}.}\label{fig:ablation}
    \end{figure}
    
    An ablation study is presented to understand the benefits of utilizing elapsed time and estimated progress as additional features in the RSD pre-training model. Figure \ref{fig:ablation} illustrates the improvement in surgical phase recognition performance of the \textit{EndoN2N} model when it is pre-trained with our proposed RSD prediction model as compared to the RSDNet model of \cite{twinanda2018rsd}. The study is performed on the first fold of the \textit{Cholec120} dataset. All 80 training videos are used for self-supervised pre-training while either 20, 40 or 80 annotated training videos are used to fine-tune \textit{EndoN2N} for surgical phase recognition. The smaller subsets of 20 and 40 videos have been sampled from the 80 training videos using the method described in section \ref{sec:pretraining_data}. It is to be noted that the vanilla \textit{EndoN2N} architecture, Figure \ref{fig:cnn_lstm}, is pre-trained when using RSDNet, while the updated model architecture, Figure \ref{fig:endon2n_updated}, is required when using our proposed RSD pre-training approach.
    
    It can clearly be seen that our proposed RSD pre-training approach leads to superior surgical phase recognition performance in terms of both accuracy and F1-Score. It is also noteworthy that when the \textit{EndoN2N} model is trained on all 80 annotated videos using the RSDNet model for pre-training (86.9\% accuracy and 80.4\% F1-score), it performs worse than the \textit{EndoN2N} model without any self-supervised pre-training  (88.2\% accuracy and 81.9\% F1-score). However, this is not the case with our proposed RSD pre-training model, which leads to an improvement in performance (89.6\% accuracy and 83.4\% F1-score) even when all annotated training data is used. 
    
    
	
	\subsection{Amount of Pre-Training Data}
    
    \begin{figure}[t]
    	\centering
        \includegraphics[width=7cm]{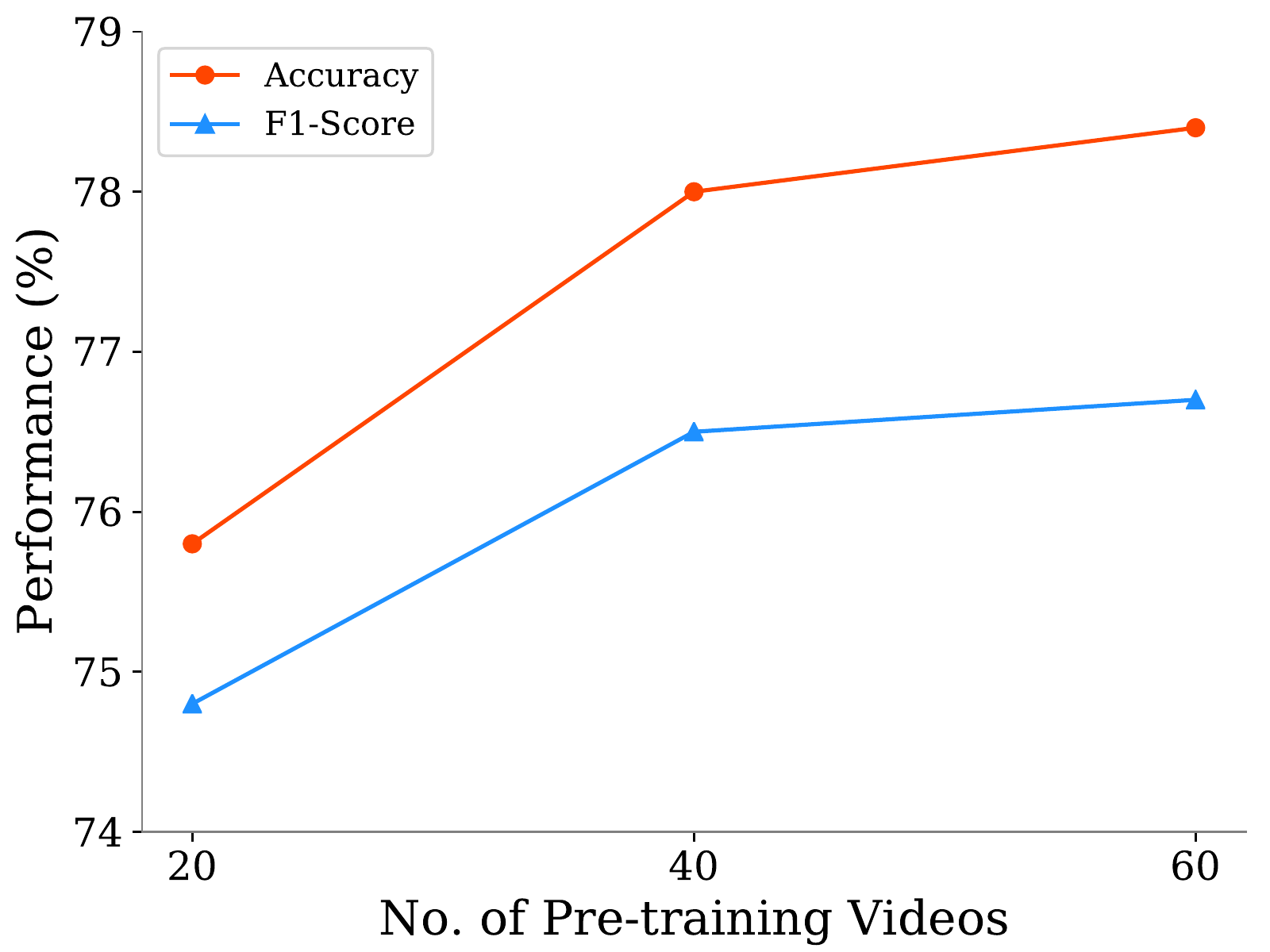}
        \caption{Graph illustrating the effect of the amount of pre-training videos utilized on surgical phase recognition performance.}\label{fig:percentage}
    \end{figure}
	
    Here, we design an experiment to study the effect of the amount of pre-training data available on our RSD pre-training approach. We first divide the 80 training videos of each fold into four quarters. 20 videos are used to fine-tune the network for surgical phase recognition. Increasing amounts of the remaining training videos, i.e., 20, 40 and 60 training videos, are used for RSD pre-training. Figure \ref{fig:percentage} shows the results of the RSD pre-trained \textit{EndoN2N} model with different amounts of pre-training videos. The results shown are the averages over the four folds. As we would intuitively expect, the accuracy and F1-score increase with greater amounts of pre-training data.
    
	\subsection{Phase Boundary Detection}
	
     \begin{figure*}[ht]
     	\begin{subfigure}[ht]{0.5\textwidth}
    	\centering
        \includegraphics[width=8cm]{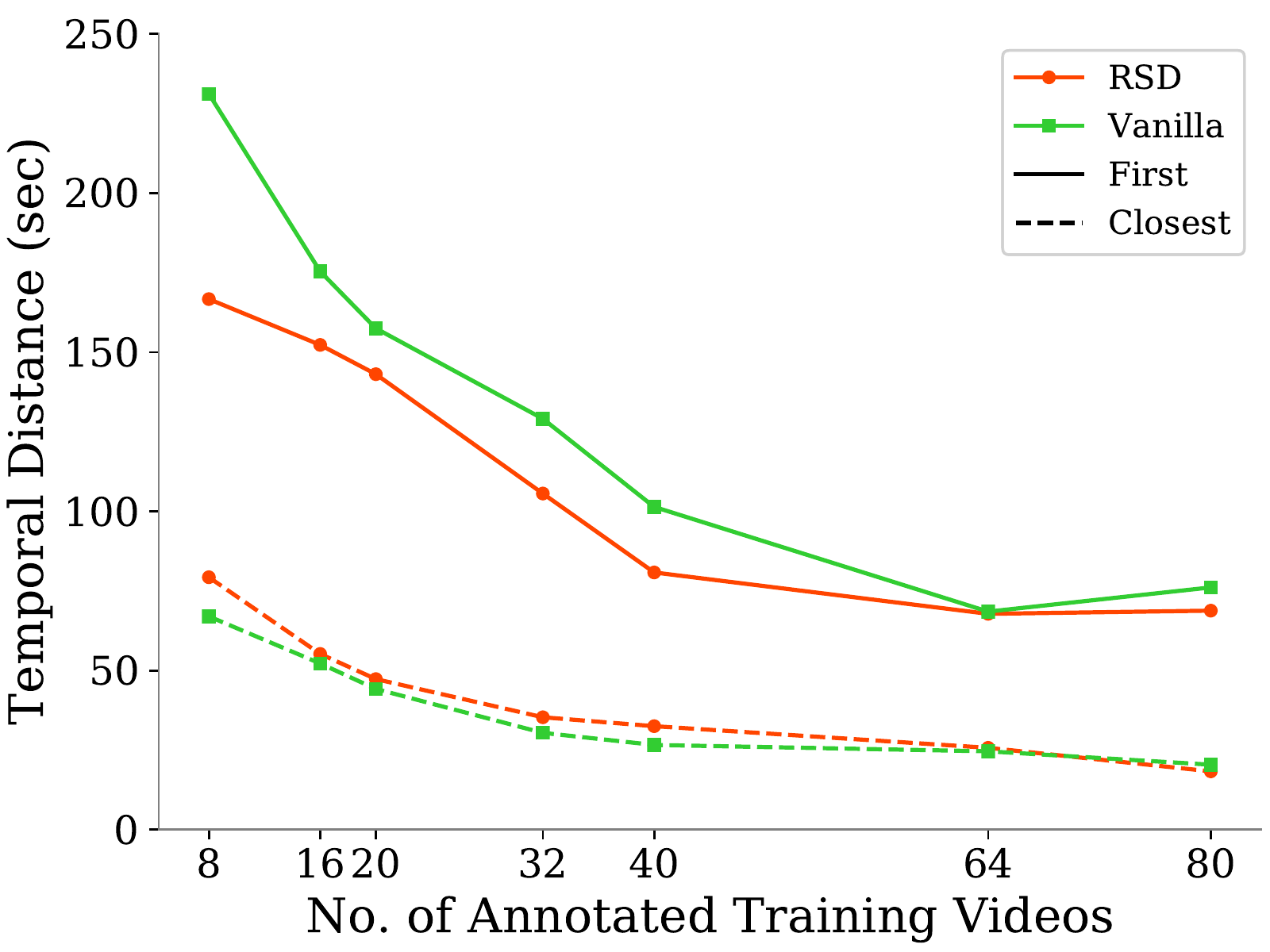}
        \caption{Temporal distance}\label{fig:temporal_distance}
        \end{subfigure}
        \begin{subfigure}[ht]{0.5\textwidth}
    	\centering
        \includegraphics[width=8cm]{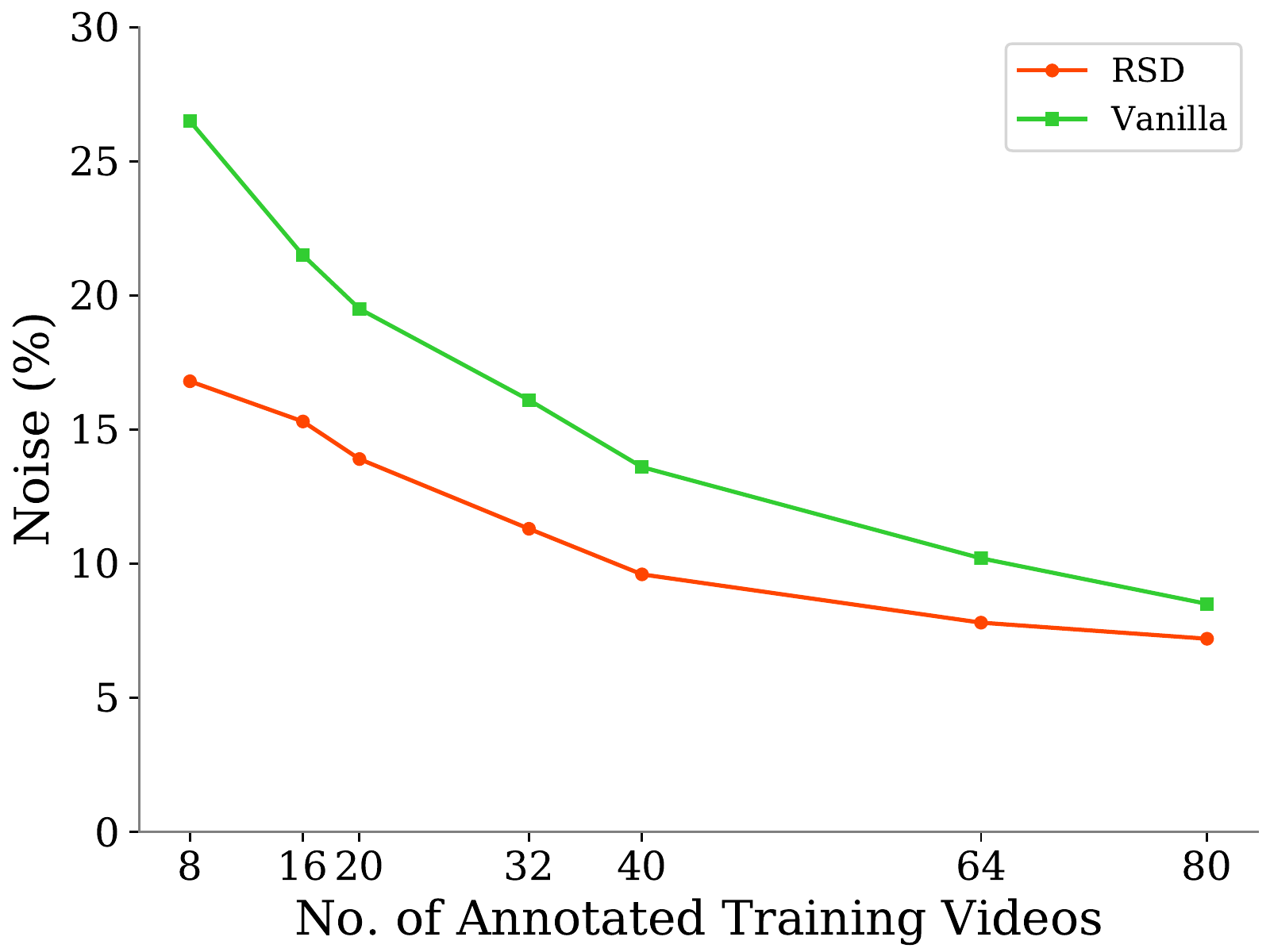}
        \caption{Noise}\label{fig:noise}
        \end{subfigure}
        \caption{Graphs depicting the variation in quality of phase boundary predictions when different amounts of annotated training videos are used, with or without the proposed RSD pre-training. (a) shows the temporal distance between the actual phase boundaries in the ground truth and the phase boundaries predicted by the RSD pre-trained and vanilla \textit{EndoN2N} models. The temporal distance is calculated with respect to both the first predicted and closest predicted phase boundaries. (b) shows the percentage of noise in the predictions.}\label{fig:boundary}
    \end{figure*}
    
    We perform an additional experiment to study how well the surgical phase recognition model is able to locate phase boundaries within a laparoscopic procedure. We measure this using the temporal distance, which is the absolute time difference in seconds between the actual phase boundary in the ground truth and the corresponding predicted phase boundary. For comparison, the temporal distance is measured with respect to both the first prediction and to the closest prediction. On the one hand, the first prediction of a phase boundary is important during actual clinical application. For example, an automatic notification system will alert the required hospital staff at the first instance a new phase is detected. On the other hand, high accuracy of the closest predicted phase boundary enables good initial annotations to be generated that can be of assistance to manual annotators. This can facilitate the creation of annotated data and be further beneficial for scaling up surgical phase recognition algorithms. 
    
    \begin{figure}[t]
    	\centering
        \includegraphics[width=8cm]{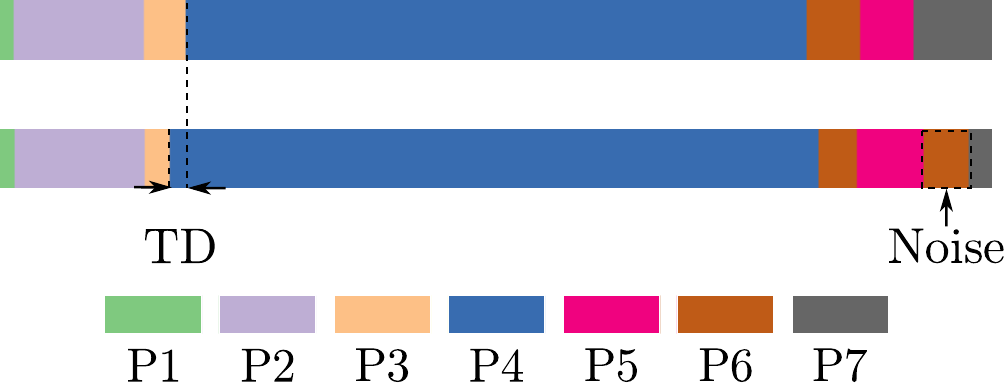}
        \caption{Visualization of ground truth (above) and RSD pre-trained \textit{EndoN2N} predictions (below) for one video of \textit{Cholec120}. The 7 color coded phase labels are displayed at the bottom. An instance of noise and temporal distance (TD) has been highlighted. (Best seen in color.)}\label{fig:predictions}
    \end{figure}
    
    Another metric we compute is the noise. Certain incorrect phase intervals are predicted by the model which do not appear in the ground truth, as shown in Figure \ref{fig:predictions}. Noise is computed as the percentage of total time steps of a laparoscopic video which belong to the incorrect phase intervals. Higher noise is detrimental both in clinical applications and when aiming to create a good set of initial annotations.
    
    The calculations presented in Figure \ref{fig:boundary} are obtained after the predictions of \textit{EndoN2N} are filtered using a 5 second window to remove any short-term noise or spikes. A 5 second delay in prediction is deemed acceptable for practical real-time applications. The results presented are the averages over all four folds of \textit{cholec120}.
    
    It can be seen from the graphs in Figure \ref{fig:boundary} that using as little as 40 annotated training videos along with RSD pre-training leads to a performance similar to the vanilla \textit{EndoN2N} model trained on all 80 annotated videos. The proposed RSD pre-training is particularly effective in improving the accuracy of the first predicted phase boundaries and reducing false predictions which contribute to noise, making it beneficial for clinical applications. The reduction in prediction noise is also beneficial for creating good initial annotations. Though the closest phase boundaries are generally more accurately predicted by the vanilla \textit{EndoN2N} model, the difference is very small (less than 5 seconds on average). It should be noted that the temporal distance is computed with respect to the phase boundaries in the ground truth, which are very strict. For practical applications, a slight error is acceptable since the phase transitions are actually more gradual. The use of a bi-directional LSTM based model could further improve predictions for creating initial annotations, although such a model can not be used in real-time applications.
    
    
\section{Conclusion}
	
    A new self-supervised pre-training approach based on RSD prediction has been presented and shown to be particularly effective in reducing the amount of annotated data required for successful surgical phase recognition. This makes our approach beneficial for scaling up surgical phase recognition algorithms to different types of surgeries. Surgical phase recognition performance when using only half the amount of annotated data generally remains within 5\% if the proposed RSD pre-training is utilized. Additionally, when a sufficiently large amount of pre-training data is utilized, surgical phase recognition performance can even be slightly improved despite relying on 20\% less annotated data. This is especially significant for real-world clinical applications, where despite a scarcity of annotated data, which is time-consuming and difficult to generate, there exists an abundance of unlabeled data. The use of self-supervised pre-training ensures that no data remains unexploited. The proposed RSD pre-training approach also outperforms the temporal context pre-training approach, the single self-supervised pre-training approach previously implemented for surgical phase recognition. Further, it is interesting to note that the proposed RSD pre-training approach leads to improvement in performance even when all the training data is annotated.
    
    This work also presents an apples-to-apples comparison between the end-to-end optimization and the two-step optimization of surgical phase recognition models based on CNN-LSTM networks. The results show that the proposed end-to-end optimization approach leads to better performance. Additional experiments were presented, which provide a greater insight into the proposed RSD pre-training model as well as the effectiveness of our models for both deployment in ORs and generation of initial surgical phase annotations.
    
    We hope this paper serves as a motivation for other works to address the important problem of developing surgical phase recognition approaches, which are less reliant on annotated data. In future work, the effectiveness of other semi-supervised approaches, such as the application of generative adversarial networks or the use of synthetic data, for example, can be explored. Additional CNN-LSTM pre-training approaches based on self-supervised learning can also prove to be effective. We would also like to carry out the proposed RSD pre-training using a much larger number of laparoscopic videos than the 80 used in this work, to study the benefit in performance that can be obtained.
    
    \section*{Acknowledgements}
    
    This work was supported by French state funds managed within the Investissements d'Avenir program by BPI France (project CONDOR) and by the ANR (references ANR-11-LABX-0004 and ANR-10-IAHU-02). The authors would also like to acknowledge the support of NVIDIA with the donation of a GPU used in this research.
    
	
	\bibliographystyle{model2-names}

\end{document}